\documentclass[lettersize,journal]{IEEEtran}
\usepackage{graphicx}
\usepackage{subfigure}
\usepackage{amsmath,amsfonts}
\usepackage{multirow}
\usepackage{comment} 
\usepackage{hhline}
\usepackage{tabularx}
\usepackage{booktabs}
\usepackage{multirow}
\usepackage{bigstrut}
\usepackage{amsmath} 
\usepackage{helvet}  
\usepackage{hyperref} 
\hypersetup{colorlinks=true,linkcolor=blue,anchorcolor=blue,citecolor=blue} 
\usepackage[table]{xcolor}
\usepackage{array}
\usepackage{algorithmic}
\usepackage{array}
\usepackage{textcomp}
\usepackage{cite}
\usepackage{stfloats}
\usepackage{url}
\usepackage{verbatim}
\usepackage{graphicx}
\usepackage{color}
\usepackage{bm}
\usepackage{mathrsfs}
\usepackage{amssymb}
\usepackage{float}
\usepackage{verbatim}
\usepackage{comment}
\usepackage{arydshln}
\def\BibTeX{{\rm B\kern-.05em{\sc i\kern-.025em b}\kern-.08em
	T\kern-.1667em\lower.7ex\hbox{E}\kern-.125emX}}
\usepackage{balance}

\begin{document}
\title{PC-NeRF: Parent-Child Neural Radiance Fields Using Sparse LiDAR Frames in Autonomous Driving Environments}
\author{Xiuzhong Hu, Guangming Xiong, Zheng Zang, Peng Jia, Yuxuan Han, Junyi Ma* 
	\thanks{
		The research is funded by the National Natural Science Foundation of China under Grant 52372404. (Corresponding author: Junyi Ma.)
		
		The authors are with the School of Mechanical Engineering, Beijing Institute of Technology, Beijing, 100081, China. (e-mail: 3120210302@bit.edu.cn; xiongguangming@bit.edu.cn; zhengzangbiter@gmail.com; xtjp960722@163.com; yx\_han\_work@foxmail.com; junyi.ma@bit.edu.cn).
		}
	}
\maketitle

\begin{abstract}
	Large-scale 3D scene reconstruction and novel view synthesis are vital for autonomous vehicles, especially utilizing temporally sparse LiDAR frames. However, conventional explicit representations remain a significant bottleneck towards representing the reconstructed and synthetic scenes at unlimited resolution. Although the recently developed neural radiance fields (NeRF) have shown compelling results in implicit representations, the problem of large-scale 3D scene reconstruction and novel view synthesis using sparse LiDAR frames remains unexplored. To bridge this gap, we propose a 3D scene reconstruction and novel view synthesis framework called parent-child neural radiance field (PC-NeRF). Based on its two modules, parent NeRF and child NeRF, the framework implements hierarchical spatial partitioning and multi-level scene representation, including scene, segment, and point levels. 	
	The multi-level scene representation enhances the efficient utilization of sparse LiDAR point cloud data and enables the rapid acquisition of an approximate volumetric scene representation.
	With extensive experiments, PC-NeRF is proven to achieve high-precision novel LiDAR view synthesis and 3D reconstruction in large-scale scenes. Moreover, PC-NeRF can effectively handle situations with sparse LiDAR frames and demonstrate high deployment efficiency with limited training epochs. Our approach implementation and the pre-trained models are available at \url{https://github.com/biter0088/pc-nerf}.
\end{abstract}


\section{Introduction}

\begin{figure}[!t]
	\centering
	\includegraphics[width=\columnwidth]{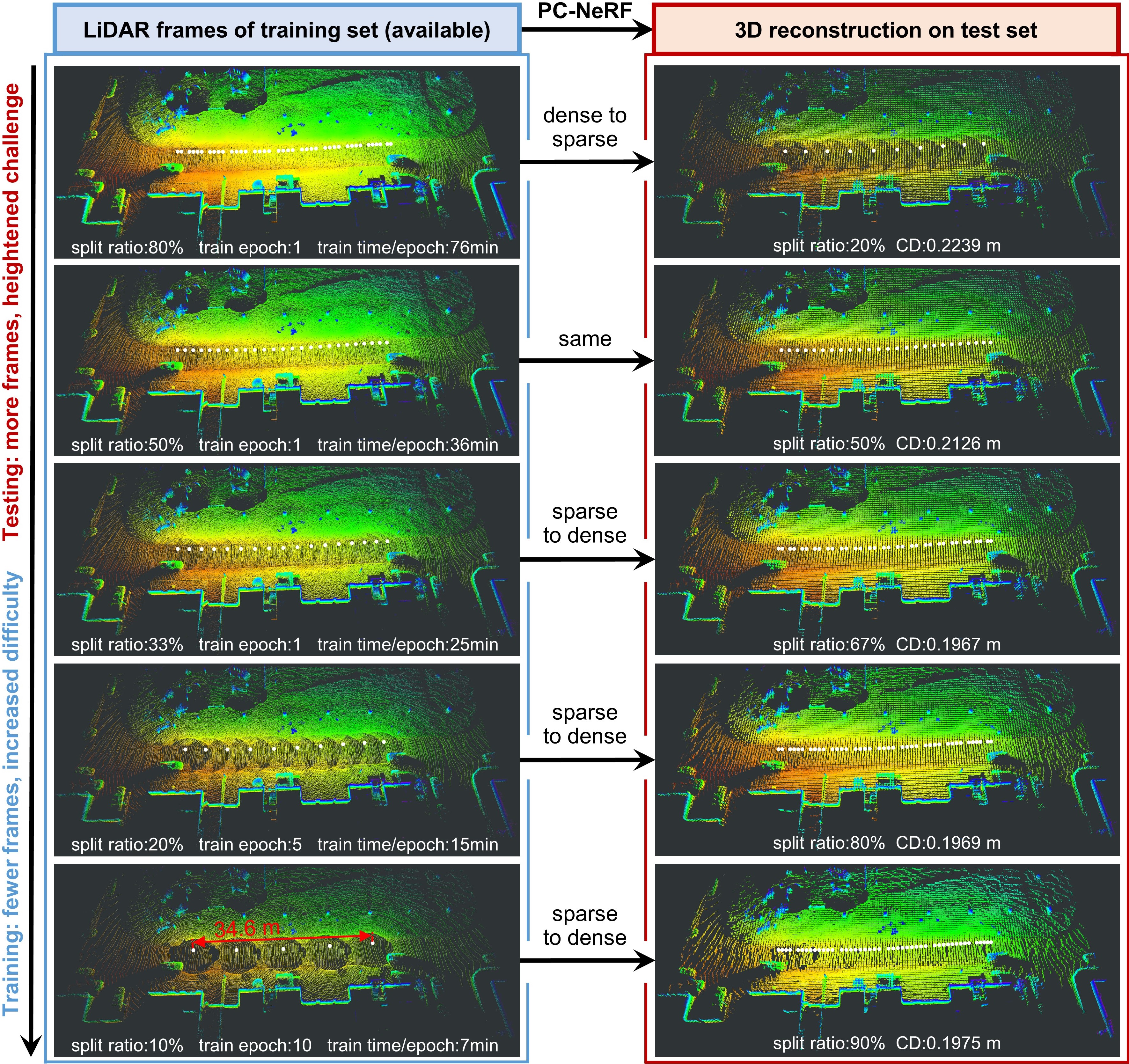}		
	\caption{PC-NeRF excels in 3D scene reconstruction and novel view synthesis, showcasing robustness to increased LiDAR frame sparsity with minimal training.
	Each subfigure depicts 3D scene reconstruction achieved by stitching real or synthetic LiDAR views with their corresponding poses. This scene involves frames 1151-1200 from the KITTI 00 sequence, encompassing diverse elements like the ground, grass, walls, and vehicles. White dots in each subfigure depict the LiDAR positions of each frame, and CD gauges 3D reconstruction accuracy, with smaller values indicating superior performance. As the proportion of LiDAR frames for training decreases, signifying increased sparsity, PC-NeRF achieves sparse-to-dense 3D reconstruction, as evident in the last three rows of subfigures. Moreover, utilizing only 33\,\% of LiDAR frames during training demonstrates advantages in both reconstruction quality and time consumption compared to using 50\,\% and 80\,\% of frames, as depicted in the first three rows of subfigures. More details are in Sec.~\ref{sec:Lesser point clouds for 3D reconstruction}.}
	\label{fig:fig_motivation2}		
\end{figure}

\IEEEPARstart{l}{arge-scale} 3D scene reconstruction and novel view synthesis are essential for autonomous vehicles to conduct environmental exploration, motion planning, and closed-loop simulation \cite{Part0, MotionPlanning, li2023point, zhong2023shine, ran2023neurar, Deng_2023_ICCV, turki2022mega, chang2022lamp}, especially when the available sensor data is temporally sparse due to various practical factors \cite{parr2023investigating, zhang2023hivegpt, wang2023data, song2023synthetic, wang2021adaptive, raouf2022sensor, waqas2022automatic}. 
Although conventional explicit representations can depict the reconstructed scene and synthetic view visually \cite{hornung2013octomap, hu2023non, vizzo2021poisson}, they remain a significant bottleneck towards representing the scene at unlimited resolution \cite{yang2023neural}, as the explicit representation is discrete. 
As a trendy method of implicit representation \cite{yu2023sketch, wang2021neus}, neural radiance fields (NeRF) \cite{mildenhall2021nerf} attract significant research interest in computer vision, robotics, autonomous driving, and augmented reality communities \cite{liu2023efficient, chen2023camera, turki2022mega, sucar2021imap, moreau2022lens, zhu2022nice, Deng_2023_ICCV, yu2023nf, tao2023lidarnerf, zhang2023nerf, Huang2023nfl, kuang2022ir, wiesmann2023locndf, rematas2022urban, tancik2022block, zhenxing2022switch, rebain2021derf, reiser2021kilonerf, liu2020neural, xiangli2022bungeenerf, martin2021nerf}. 
NeRF typically represents a scene using a fully connected deep network, which maps a single continuous 5D coordinate (spatial location and viewing direction) to the volume density and view-dependent emitted radiance at that spatial location \cite{mildenhall2021nerf}. 
Hence, NeRF can construct a smooth, continuous, and differentiable scene representation \cite{chen2023camera}, which helps utilize as much available sparse sensor data as possible. 
Most NeRF-related works have been carried out based on camera image data or indoor laser scan data \cite{yang2023neural, sucar2021imap, moreau2022lens, zhu2022nice, turki2022mega, tancik2022block, zhenxing2022switch, rebain2021derf, reiser2021kilonerf, liu2020neural, wang2021neus, liu2023efficient, kuang2022ir, martin2021nerf, xiangli2022bungeenerf}, with only a few works using LiDAR point cloud data in large-scale outdoor environments \cite{Deng_2023_ICCV, tao2023lidarnerf, zhang2023nerf, Huang2023nfl}. 
Unlike cameras, LiDAR has the capability to directly capture accurate distance information \cite{ma2022overlaptransformer, wang2022performance}, making it particularly valuable for high-precision outdoor mapping in large-scale environments and dealing with complex scene geometry in NeRF \cite{yu2023nf, Deng_2023_ICCV}.
However, to leverage the successes of image-based NeRF methods, many LiDAR-based NeRF approaches project 3D LiDAR point clouds onto 2D range pseudo-images, resulting in substantial information loss \cite{zhang2023nerf, tao2023lidarnerf, milioto2019rangenet++}. 
Besides, in real autonomous vehicle applications, some unfavorable conditions, such as hardware failures and unstable communication in remote control tasks, may aggravate the temporal sparsity of LiDAR frames due to missing observation \cite{parr2023investigating, zhang2023hivegpt, wang2023data, song2023synthetic, wang2021adaptive, raouf2022sensor, waqas2022automatic}. 
Therefore, exploring NeRF-based methods for effective utilization of sparse LiDAR frames is crucial to enhance vehicles' autonomous capabilities, particularly evident in 3D scene reconstruction and novel view synthesis as depicted in Fig.~\ref{fig:fig_motivation2}.

This paper presents a parent-child neural radiance fields (PC-NeRF) framework for large-scale 3D scene reconstruction and novel LiDAR view synthesis optimized for efficiently utilizing sparse LiDAR frames in outdoor autonomous driving.
	PC-NeRF incorporates a hierarchical spatial partitioning approach and a multi-level scene representation. 
	The hierarchical spatial partitioning approach first divides the driving environment into multiple large blocks, labelled as parent NeRFs, and then further partitions each parent NeRF by extracting child NeRFs.
	Unlike the collected LiDAR point clouds, which are often sparse and do not completely cover the object surfaces, each bounding-box-wise child NeRF represents a point cloud segment, encompassing a collection of closely located laser points and its surrounding area. 
	The parent NeRF shares the network with child NeRFs within it for implicitly unified spatial representations. 
	Based on the hierarchical spatial partitioning approach, we propose a multi-level scene representation, including scene, segment, and point levels.
	Compared to scene-level and point-level representations that represent the scene in its entirety and detail, respectively, segment-level representations try to represent the individual objects within the scene.
	Recognizing the inherent limitation of LiDAR point clouds in providing discrete samples of actual object surfaces, we choose the segment-level representation over the ideal object-level one. 
	This choice facilitates the swift capture of the approximate object distribution in the environment, even in the presence of sparse LiDAR frames.

To sum up, the primary contributions of this paper include:

$\bullet$ To our knowledge, our proposed PC-NeRF is the first NeRF-based large-scale 3D scene reconstruction and novel LiDAR view synthesis method using sparse LiDAR frames, even though NeRF is a dense volumetric representation typically constructed using large amounts of sensor data. 

$\bullet$ 
To represent outdoor large-scale autonomous driving environments, PC-NeRF introduces a hierarchical spatial partitioning approach, progressively dividing the driving environment into parent and child NeRFs.

$\bullet$ 
Based on the hierarchical spatial partitioning approach, we propose a multi-level scene representation to optimize scene-level, segment-level, and point-level representations concurrently.
The multi-level scene representation is capable of efficiently utilizing sparse LiDAR frames, along with achieving high-precision 3D scene reconstruction and novel LiDAR view synthesis with minimal training epochs.

\section{Related Works}
With NeRF's inherent advantages of continuous dense volumetric representation, NeRF-based techniques in novel view synthesis \cite{tao2023lidarnerf, zhang2023nerf, Huang2023nfl, liu2020neural, rebain2021derf, reiser2021kilonerf}, scene reconstruction \cite{rematas2022urban, turki2022mega, tancik2022block, zhenxing2022switch, xiangli2022bungeenerf, liu2020neural, rebain2021derf, reiser2021kilonerf}, and localization systems \cite{sucar2021imap, moreau2022lens, zhu2022nice, Deng_2023_ICCV, kuang2022ir, wiesmann2023locndf, yang2022vox} have rapidly developed and are highly referential and informative. 
PC-NeRF utilizes NeRF's capability for continuous scene representation modelling, employs LiDAR data as inputs, and hierarchically divides the scene spaces.
Therefore, this section reviews the literature on LiDAR-based NeRF and space-division-based NeRF.

\subsection{LiDAR-Based NeRF}
Motivated by NeRF's capability to render photo-realistic novel image views, several studies have investigated its potential application to LiDAR point cloud data
for novel view synthesis \cite{tao2023lidarnerf, zhang2023nerf, Huang2023nfl} and robot navigation \cite{Deng_2023_ICCV, kuang2022ir, wiesmann2023locndf}.

The goal of novel view synthesis is to generate a view of a 3D scene from a viewpoint where no real sensor image has been captured, providing the opportunity to observe real scenes from a virtual perspective. Neural Fields for LiDAR (NFL) \cite{Huang2023nfl} combines the rendering power of neural fields with a physically motivated model of the LiDAR sensing process, thus enabling it to accurately reproduce key sensor behaviors like beam divergence, secondary returns, and ray drop. Our work is inspired by the modeling of LiDAR sensing processes. LiDAR-NeRF \cite{tao2023lidarnerf} converts the 3D point cloud into the range pseudo image in 2D coordinates and then optimizes three losses, including absolute geometric error, point distribution similarity, and realism of point attributes. Similar to LiDAR-NeRF, NeRF-LiDAR \cite{zhang2023nerf} has also employed the spherical projection strategy and consists of three key components: NeRF reconstruction of the driving scenes, realistic LiDAR point clouds generation, and point-wise semantic label generation. 
However, when multiple laser points project onto the same pseudo-pixel, only the one with the smallest distance is retained \cite{tao2023lidarnerf}. This effect becomes particularly pronounced with small resolution range pseudo-images, leading to significant information loss in cases of spherical projections \cite{milioto2019rangenet++}.

In robotics, LiDAR-based NeRF is usually proposed for localization and mapping. IR-MCL \cite{kuang2022ir} focuses on the problem of estimating the robot's pose in an indoor environment using 2D LiDAR data. With the pre-trained network, IR-MCL can synthesize 2D LiDAR scans for an arbitrary robot pose through volume rendering. However, the error between the synthesized and real scans is relatively large. NeRF-LOAM \cite{Deng_2023_ICCV} presents a novel approach for simultaneous odometry and mapping using neural implicit representation with 3D LiDAR data. NeRF-LOAM employs sparse octree-based voxels combined with neural implicit embeddings, decoded into a continuous signed distance function (SDF) by a neural implicit decoder. However, NeRF-LOAM cannot currently operate in real-time with its unoptimized Python implementation. LocNDF \cite{wiesmann2023locndf} utilizes neural distance fields (NDFs) for robot localization, demonstrating the direct learning of NDFs from range sensor observations. LocNDF has raised the challenge of addressing real-time constraints, and our work endeavors to investigate this challenge.

In contrast to projecting the LiDAR point cloud onto a range pseudo-image, our proposed PC-NeRF handles 3D LiDAR point cloud data directly. Besides learning the LiDAR beam emitting process, our proposed PC-NeRF explores the deployment performance of NeRF-based methods.

\subsection{Space-Division-Based NeRF}
When large-scale scenes such as where the autonomous vehicles drive need to be represented with high precision, the model capacity of a single NeRF is limited in capturing local details with acceptable computational complexity \cite{turki2022mega, tancik2022block, zhenxing2022switch}. For large-scale 3D scene reconstruction tasks, Mega-NeRF \cite{turki2022mega}, Block-NeRF \cite{tancik2022block}, and Switch-NeRF \cite{zhenxing2022switch} have adopted the multiple NeRF solution, with each NeRF responsible for different scene areas. Mega-NeRF decomposes a scene into cells with centroids and initializes a corresponding set of model weights. At query time, Mega-NeRF produces opacity and color for a given position and direction using the model weights closest to the query point. Like Mega-NeRF, Block-NeRF \cite{tancik2022block} proposes dividing large environments into individually trained Block-NeRFs, which are then rendered and combined dynamically at inference time. For rendering a target view, a subset of the Block-NeRFs are rendered and then composited based on their geographic location compared to the camera. Switch-NeRF \cite{zhenxing2022switch} proposes a novel end-to-end large-scale NeRF with learning-based scene decomposition and designs a gating network to dispatch 3D points to different NeRF sub-networks. The gating network can be optimized with the NeRF sub-networks for different scene partitions by design with the Sparsely Gated Mixture of Experts. Besides the multiple NeRF solution, NeRF-LOAM and Shine-mapping employ the octree structure to recursively divide the scene into leaf nodes with basic scene unit voxels, simplifying the description of large-scale scenes \cite{Deng_2023_ICCV, zhong2023shine, liu2023efficient}. These basic scene unit voxels attach an N-dimensional encoding at each vertex and share it with neighboring voxels. Thus, the attributes of any 3D location in the scene can be inferred from the vertex encoding values output by the neural network, which in turn achieves 3D reconstruction.

Regarding space-division-based NeRF on a smaller scale, further space division is employed for faster and higher-quality rendering \cite{rebain2021derf, reiser2021kilonerf, liu2020neural}. DeRF \cite{rebain2021derf} and KiloNeRF \cite{reiser2021kilonerf} adopt the multiple NeRF solution to represent scene details and speed up rendering. DeRF \cite{rebain2021derf} decomposes the scene space and represents each decomposed part using a separate neural network (also named a decomposition head network), where each decomposition head network is defined over the entire scene space. Using Voronoi learnable decompositions, only one of the decomposition head networks works at any given spatial location. Thus, only one decomposition head network needs to be evaluated for each spatial location, resulting in an accelerated inference process. KiloNeRF \cite{reiser2021kilonerf} demonstrates that real-time rendering is possible using thousands of tiny MLPs instead of one extensive multilayer perceptron network (MLP). Rather than representing the entire scene with a single, high-capacity MLP, KiloNeRF represents the scene with thousands of small MLPs. Similar to the octree scene partition on a large-scale scene, Neural Sparse Voxel Fields (NSVF) \cite{liu2020neural} also partitions the scene space with the sparse voxel octree and assigns the voxel embedding at each vertex.

Our work introduces a hierarchical spatial partitioning approach that incorporates spatial division concepts at both large and small scales for representing autonomous driving scenes.
We first partition the driving environment into multiple large blocks and then, within each block, extract the spatial extents of point cloud segments.
We also employ the multiple NeRF solution, where the volumetric representation of a large block and the point cloud segments inside it are represented by a shared NeRF network.
Unlike the octree scene partition and voxel vertex embedding, we concentrate on constructing the volumetric representation around and within individual point cloud segments, as the point cloud segments can represent the approximate spatial extent of the actual object surfaces and thus alleviate the sparsity of the LiDAR frames. 
Therefore, we opt for a segment-level representation instead of the ideal object-level representation to rapidly access to the detailed structural features of a scene.

\begin{figure*}[t]
	\centering
	\includegraphics[width=7.0in]{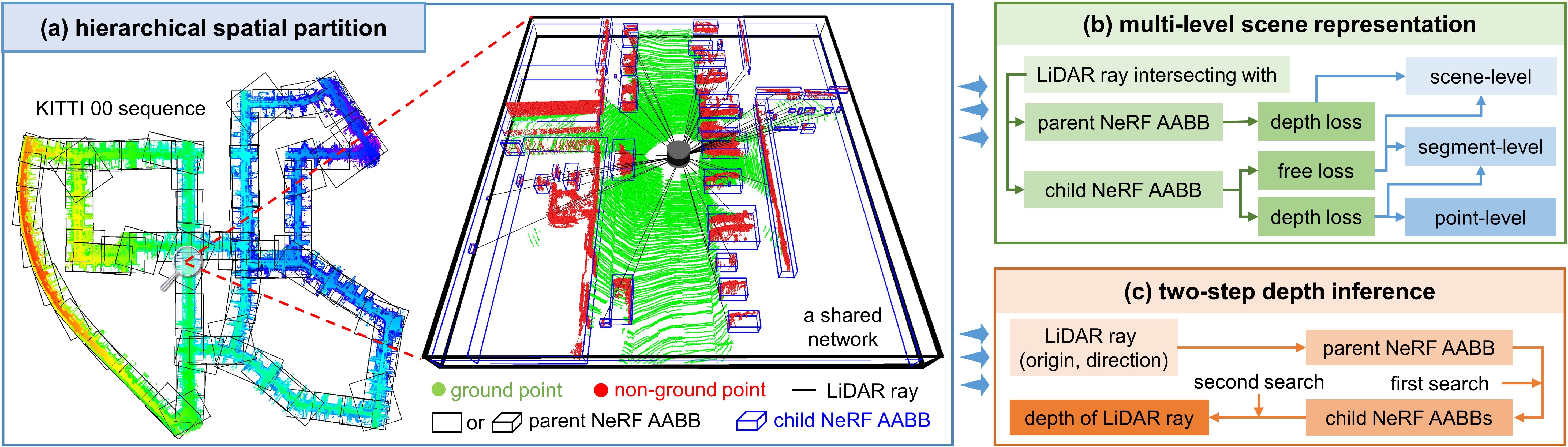}%
	\caption{Our PC-NeRF framework: (a) The hierarchical spatial partition divides the entire large-scale scene into large blocks, referred to as parent NeRFs. After multi-frame point cloud fusing, ground filtering, and non-ground point cloud clustering, a large block is further divided into point cloud geometric segments represented by a child NeRF.
	The parent NeRF shares a network with the child NeRFs within it. (b) In the multi-level scene representation, the surface intersections of the LiDAR ray with the parent and child NeRF AABBs and the LiDAR origin are used to divide the entire LiDAR ray into different line segments. The three losses on these line segments concurrently optimize the scene representation at the scene level, segment level, and point level, effectively leveraging sparse LiDAR frames. (c) For depth inference of each LiDAR ray, PC-NeRF searches in the parent NeRF AABB to locate corresponding child NeRF AABBs and then refines its inference in the child NeRF AABBs for higher precision.}
	\label{framework}
\end{figure*}

\section{Parent-Child Neural Radiance Fields (PC-NeRF) Framework}
To explore the large-scale 3D scene reconstruction and novel LiDAR view synthesis based on LiDAR and NeRF, especially using sparse LiDAR frames, we propose a parent-child neural radiance field (PC-NeRF) framework, as shown in Fig.~\ref{framework}.
We introduce a hierarchical spatial partitioning approach in Sec.~\ref{sec:parent NeRF} and Sec.~\ref{sec:Child NeRFs}, dividing the entire autonomous vehicle driving environment into large blocks, i.e., parent NeRFs, and further dividing a block into geometric segments, represented by child NeRFs. A parent NeRF shares a network with child NeRFs within it. 
Based on the hierarchical spatial partitioning approach, Sec.~\ref{sec:LiDAR loss} introduces a comprehensive multi-level scene representation aimed at collectively optimizing scene-level, segment-level, and point-level representations while efficiently utilizing sparse LiDAR frames.
Moreover, we propose a two-step depth inference method in Sec.~\ref{sec:Two-step Range Value Inference} to realize segment-to-point inference.

\subsection{Spatial Partition of Parent NeRF}\label{sec:parent NeRF}
To efficiently represent the autonomous vehicle's driving environment, our proposed hierarchical spatial partitioning approach constructs multiple rectangular-shaped parent NeRFs (as Fig.~\ref{framework}(a) illustrates) one after the other along the trajectory. A new parent NeRF is created when the autonomous vehicle orientation variation exceeds a given threshold. Besides, parent NeRFs with highly overlapping spatial areas should be merged. 
Constructing parent NeRF requires three considerations: the effective LiDAR point cloud selection, the driving environment representation, and the NeRF near/far bounds calculation. 
Since the point clouds become sparser the further away from the LiDAR origin and the NeRF itself is a dense volumetric representation, LiDAR points within a certain distance from the LiDAR origin rather than the holistic point clouds are chosen to train our proposed PC-NeRF model. 
Moreover, the driving environments featured with roads, walls, and vehicles can be tightly enclosed by bounding boxes. Therefore, each parent NeRF's space is represented as a large Axis-Aligned Bounding Box (AABB) in our work, making it easier to calculate the related near and far bounds for rendering.

\subsection{Spatial Partition of Child NeRF}\label{sec:Child NeRFs}
To efficiently capture the approximate environmental distributions using sparse LiDAR frames, our proposed hierarchical spatial partitioning approach distributes the point clouds in a parent NeRF into multiple child NeRFs (as Fig.~\ref{framework}(a) illustrates).
In contrast to the collected LiDAR point clouds, which are often sparse and do not completely cover the object surfaces, each child NeRF represents a point cloud segment, encompassing a collection of laser points close to each other and its surrounding area.
Since the geometric point cloud segment quantity is limited and the child NeRFs' space can also be represented by AABBs, constructing child NeRFs from the raw point cloud is a fast way to generate detailed environmental representations. With less total space volume, child NeRFs can represent a larger environment with the same model capacity.
Moreover, by understanding the spatial distribution of point cloud segments, child NeRFs can address environmental representation inadequacies due to the sparsity of LiDAR frames.

The point cloud allocation for child NeRF can be divided into the following three steps, and the results are shown in Fig.~\ref{framework}(a). Step 1: 
Extract ground point clouds, such as road surfaces and sidewalks along roads, from the fused point cloud data.
The ground plane is one significant geometric segment in the driving environment and is spatially adjacent to the other individual geometric segments. Therefore, distinguishing it helps to extract other geometric segments further accurately. Step 2: Cluster the remaining non-ground point clouds into various segments using region-growing clustering. The advantages of the utilized clustering method are excellent scalability for operation on large-scale point clouds and the adaptive ability to cluster different shaped and sized objects. Step 3: Construct child NeRF AABBs. The AABBs of the segmented point clouds obtained from step 1 and step 2 can be used as child NeRFs' spatial extent. After the division process above, the fused point clouds are distributed into a limited quantity of child NeRFs. Note that when the driving scene is complex, there are many artifacts existing in the results of ground extraction and point cloud clustering. However, our proposed method mitigates the negative effects of the inherent inaccuracy for these two tasks in complicated driving scenes, which is further introduced in Sec.~\ref{sec:LiDAR loss}.

\subsection{Multi-level Scene Representation for Sparse LiDAR Frames}\label{sec:LiDAR loss}
\begin{figure}[H]
	\centering
	\includegraphics[width=3.45in]{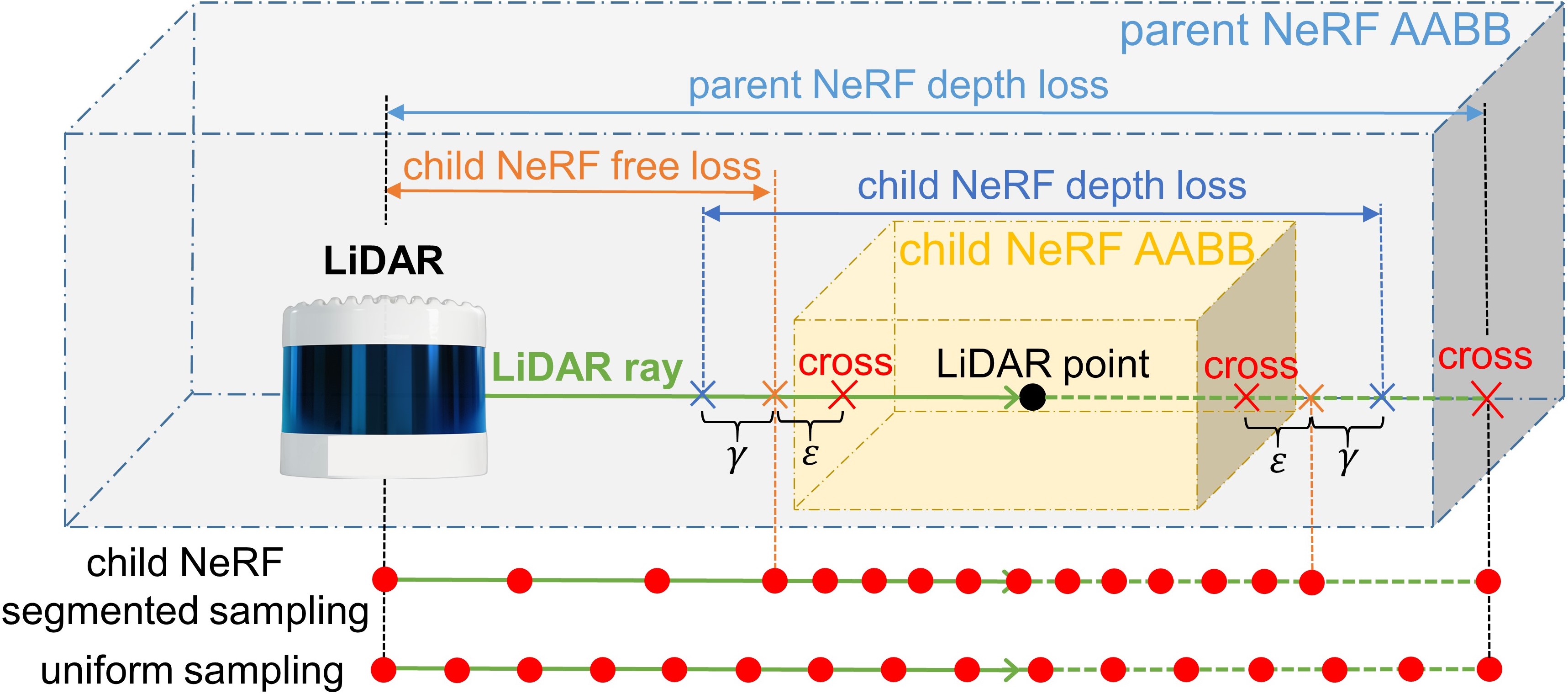}%
	\caption{Three LiDAR losses and child NeRF segmented sampling. The three LiDAR losses include parent NeRF depth loss, child NeRF depth loss, and child NeRF free loss. Using different sampling densities, the Child NeRF segmented sampling uniformly samples both inside and outside the intersection of the LiDAR ray with the Child NeRF.}
	\label{fig3}
\end{figure}

By observing the intersections of the LiDAR rays with parent and child NeRF AABBs (Fig.~\ref{framework}(a) and Fig.~\ref{fig3}), we propose a multi-level scene representation (Fig.~\ref{framework}(b)), including scene-level, segment-level, and point-level representations. The multi-level scene representation enables the simultaneous extraction of global, local, and detailed information from the environment, enhancing the utilization of effective information in sparse LiDAR frames. 
When the LiDAR ray intersects with parent and child NeRF AABBs, we can obtain three intersecting positions of the LiDAR ray with child NeRF AABB’s inner surface, child NeRF AABB’s outer surface, and parent NeRF AABB’s outer surface, as shown in and Fig.~\ref{fig3}.
The LiDAR origin and these three intersections divide the LiDAR ray into multiple line segments. 
By examining the distribution of object surface points on these line segments, we can derive the positions of individual object surface points (point-level representation), the spatial distribution of point segments formed by these object surface points (segment-level representation), and the overall distribution of object surface points in the whole space (scene-level representation).

In the global coordinate system, the LiDAR origin position of the $i$-th frame is represented as $\mathbf{o}_{i} = (x_i, y_i, z_i)$. The depth, direction, and the corresponding child NeRF order number of the $j$-th laser point $\mathbf{p}_{ij} = (x_{ij}, y_{ij}, z_{ij})$ in the $i$-th frame point cloud are $d_{ij} = \|(\mathbf{p}_{ij}-\mathbf{o}_{i})\|_2$, $\mathbf{d}_{ij} = (\mathbf{p}_{ij}-\mathbf{o}_{i})/d_{ij}$ and $k_{ij}$ respectively. Then, we can get one LiDAR ray $\mathbf{r}_{ij} = [\mathbf{o}_{i}, \mathbf{p}_{ij}, d_{ij}, \mathbf{d}_{ij}, k_{ij}]$. We apply LiDAR point cloud data to train our proposed PC-NeRF with model parameters $\bm{\theta}$. 
Besides, we further design three LiDAR losses and child NeRF segmented sampling, illustrated in Fig.~\ref{fig3}.

In NeRF volume rendering, the depth value of a ray $\mathbf{r}$ is synthesized from the weighted sampling depth values $t$ between the near and far bounds ($t_\mathrm{n}$ and $t_\mathrm{f}$) by:
\begin{equation}
	\label{equation2}
	d(\mathbf{r}) = \int_{t_\mathrm{n}}^{t_\mathrm{f}} w(t) \cdot t dt	
\end{equation}
where $\mathbf{r}(t) = \mathbf{o} + t\mathbf{d}$ represents a ray with camera or LiDAR origin $\mathbf{o}$ oriented as $\mathbf{d}$, and the volume rendering integration weights $w(t)$ is calculated by:
\begin{equation}
	\label{equation3}
	w(t) = \exp\left(-\int_{t_\mathrm{n}}^{t} \sigma(s)ds\right) \cdot \sigma(t)
\end{equation}
where $\exp(-\int_{t_n}^{t} \sigma(s)ds)$ is the visibility of $\mathbf{r}(t)$ from origin $\mathbf{o}$, and $\sigma(t)$ is the volumetric density at $\mathbf{r}(t)$.

\textbf{Parent NeRF depth loss}: In Sec.~\ref{sec:parent NeRF}, the autonomous vehicle driving environment is represented efficiently using parent NeRF. We use the intersection of the LiDAR ray $\mathbf{r}_{ij}$ with its corresponding parent NeRF surface as the far bound $f_{ij}^{p}$, as seen in Fig.~\ref{fig3}. To adequately represent the volumetric distribution of the whole driving environment, we set parent NeRF depth loss as:
\begin{equation}
	\label{equation7_1}
	\begin{aligned}
		\mathcal{L}_{ij}^{\mathrm{pd}}(\bm{\theta}) = 
		\mathcal{L}_{\mathrm{L1}}^{'}\left (\int_{t_0}^{f_{ij}^{p}} w(t) \cdot tdt, d_{ij}\right )		
	\end{aligned}
\end{equation}	
where the integration lower limit $t_0$ is set to 0. Considering the space occupied by the LiDAR or the autonomous vehicle, setting $t_0$ to $0.5\,m$ or $1\,m$ is also recommended. Moreover, $\mathcal{L}_{\mathrm{L1}}^{'}(x, y) = 0.1\cdot \text{SmoothL1Loss}(10\cdot x, 10\cdot y)$ is an extension of $\text{SmoothL1Loss}$ that shifts the $|x-y|$ turning point from $1\,m$ to $0.1\,m$ to improve model sensitivity.

The commonly used depth inference approach uses the synthetic depth between the scene's near and far bounds as the inference depth \cite{mildenhall2021nerf, kuang2022ir, rematas2022urban, sucar2021imap, zhu2022nice}, which is used in Eq.~\ref{equation7_1} and calculated as follows:
\begin{equation}
	\label{equation7_2}
	\begin{aligned}
		\hat{d_{ij}^1} = \int_{t_0}^{f_{ij}^{p}} w(t) \cdot tdt	
	\end{aligned}
\end{equation}	

\textbf{Segment-to-point hierarchical representation strategy}:
To find object surface points effectively and address the sparsity of LiDAR frames, we further propose a segment-to-point hierarchical representation strategy. The depth difference $f_{ij}^{p} - t_0$ between parent NeRF near and far bounds in large-scale outdoor scenes can be about ten times greater than that in indoor scenes, making it more difficult for the NeRF model to find the object surface through sampling. Given that it is much easier to find segment-level child NeRF space than to find point-level object surface points between the parent NeRF near and far bounds, we propose a segment-to-point hierarchical representation strategy, which first finds the child NeRFs that intersect the LiDAR ray and then finds object surface points inside the child NeRFs that intersect the LiDAR ray.

We use the intersections of the LiDAR ray $\mathbf{r}_{ij}$ with its corresponding child NeRF surface as the child NeRF near and far bounds $[n_{ij}^{c}, f_{ij}^{c}]$, as seen in Fig.~\ref{fig3}. It is pretty evident that the depth difference between child NeRF near and far bounds $(f_{ij}^{c} - n_{ij}^{c})$ is much smaller than that between parent NeRF near and far bounds. Considering that the object surface has a certain thickness and the object surface may appear at the child NeRF bounds, we slightly inflate the child NeRF near and far bounds as $[n_{ij}^{c} - \varepsilon, f_{ij}^{c}+ \varepsilon]$, where $\varepsilon $ is a small inflation coefficient, as shown in Fig.~\ref{fig3}.

\textbf{Child NeRF free loss}: To make the process of finding the segment-level child NeRF's corresponding space faster, we propose the child NeRF free loss in Eq.~\ref{equation5_1}. As no opaque objects exist in $(t_0, n_{ij}^{c} - \varepsilon)$ and no objects can be observed in $(f_{ij}^{c}+ \varepsilon, f_{ij}^{p})$ in numerous cases, the loss is calculated by:
\begin{equation}
	\label{equation5_1}
	\begin{aligned}
		\mathcal{L}_{ij}^{\mathrm{cf}}(\bm{\theta}) = \int_{t_0}^{n_{ij}^{c} - \varepsilon}w(t)^2dt + \int_{f_{ij}^{c}+ \varepsilon}^{f_{ij}^{p}} w(t)^2dt
	\end{aligned}
\end{equation}

\textbf{Child NeRF depth loss}: Based on child NeRF free loss, we propose child NeRF depth loss to find object surface points inside the child NeRF, which is calculated by:

\begin{equation}
	\label{equation5}
	\begin{aligned}
		&\mathcal{L}_{ij}^{\mathrm{cd}}(\bm{\theta}) = 
		&\mathcal{L}_{\mathrm{L1}}^{'}\left (\int_{n_{ij}^{c} - \varepsilon}^{f_{ij}^{c}+ \varepsilon} w(t) \cdot tdt, d_{ij}\right )
	\end{aligned}
\end{equation}

To let the child NeRF free loss and the child NeRF depth loss have a smooth transition at the child NeRF bounds, $\mathcal{L}_{ij}^{\mathrm{cd}}(\bm{\theta})$ is further modified to:
\begin{equation}
	\label{equation5_2}
	\begin{aligned}
		&\mathcal{L}_{ij}^{\mathrm{cd}}(\bm{\theta}) = 
		&\mathcal{L}_{\mathrm{L1}}^{'}\left (\int_{n_{ij}^{c} - \varepsilon - \gamma}^{f_{ij}^{c}+ \varepsilon +\gamma} w(t) \cdot tdt, d_{ij}\right )
	\end{aligned}
\end{equation}
where $\gamma$ is a constant designed to represent the smooth transition interval on a LiDAR ray between the child NeRF free loss and the child NeRF depth loss, as seen in Fig.~\ref{fig3}. Child NeRF free loss and child NeRF depth loss are employed separately to supervise the space from the LiDAR origin to the vicinity of the child NeRF AABB and the spatial extent of the child NeRF AABB after it expands $\gamma$ (in m). For the transition from free space to the object’s surface, the child NeRF free loss is stringent, while the child NeRF depth loss is comparatively lenient. Both functions collaboratively contribute to accurately capturing the real surface of the object. Meanwhile, they help mitigate the impact of incomplete ground extraction and insufficient point cloud clustering discussed in Sec.~\ref{sec:Child NeRFs}.

To sum up, the total training loss from one LiDAR ray $\mathbf{r}_{ij}$ contains all the losses mentioned above:
\begin{equation}
	\label{equation8}
	\begin{aligned}
		\mathcal{L}_{ij}(\bm{\theta}) =  \lambda_{\mathrm{pd}}\mathcal{L}_{ij}^{\mathrm{pd}}(\bm{\theta}) + \lambda_{\mathrm{cf}} \mathcal{L}_{ij}^{\mathrm{cf}}(\bm{\theta}) + 
		\lambda_{\mathrm{cd}}\mathcal{L}_{ij}^{\mathrm{cd}}(\bm{\theta}) 
	\end{aligned}	
\end{equation}
where $\lambda_{\mathrm{pd}}$, $\lambda_{\mathrm{cf}}$, and $\lambda_{\mathrm{cd}}$ are the parameters to jointly optimize different losses.

\textbf{Child NeRF segmented sampling}: To efficiently find the objects in large-scale scenes, even using sparse LiDAR frames, we propose a child NeRF segmented sampling method for objects more likely to be found in and around the child NeRFs. Assuming that $N$ points are sampled uniformly along the LiDAR rays, the child NeRF segmented sampling is sampling $\lambda_{\mathrm{in}} \cdot  N$ points in $[n_{ij}^{c} - \varepsilon, f_{ij}^{c} + \varepsilon] $ and sampling $( 1-\lambda_{\mathrm{in}}) \cdot  N$ in $[t_0, f_{ij}^{p}]$, as shown in Fig.~\ref{fig3}. Therefore, child NeRF segmented sampling guarantees that at least $\lambda_{\mathrm{in}} \cdot  N$ points are sampled inside child NeRF, which means that at least $\lambda_{\mathrm{in}} \cdot  N$ sampling points are sampled near the real object.

\subsection{Two-step Depth Inference}\label{sec:Two-step Range Value Inference}
\begin{figure}[!t]
	\centering
	\subfigure[LiDAR rays intersect with different AABBs.]{\includegraphics[width=3.2in]{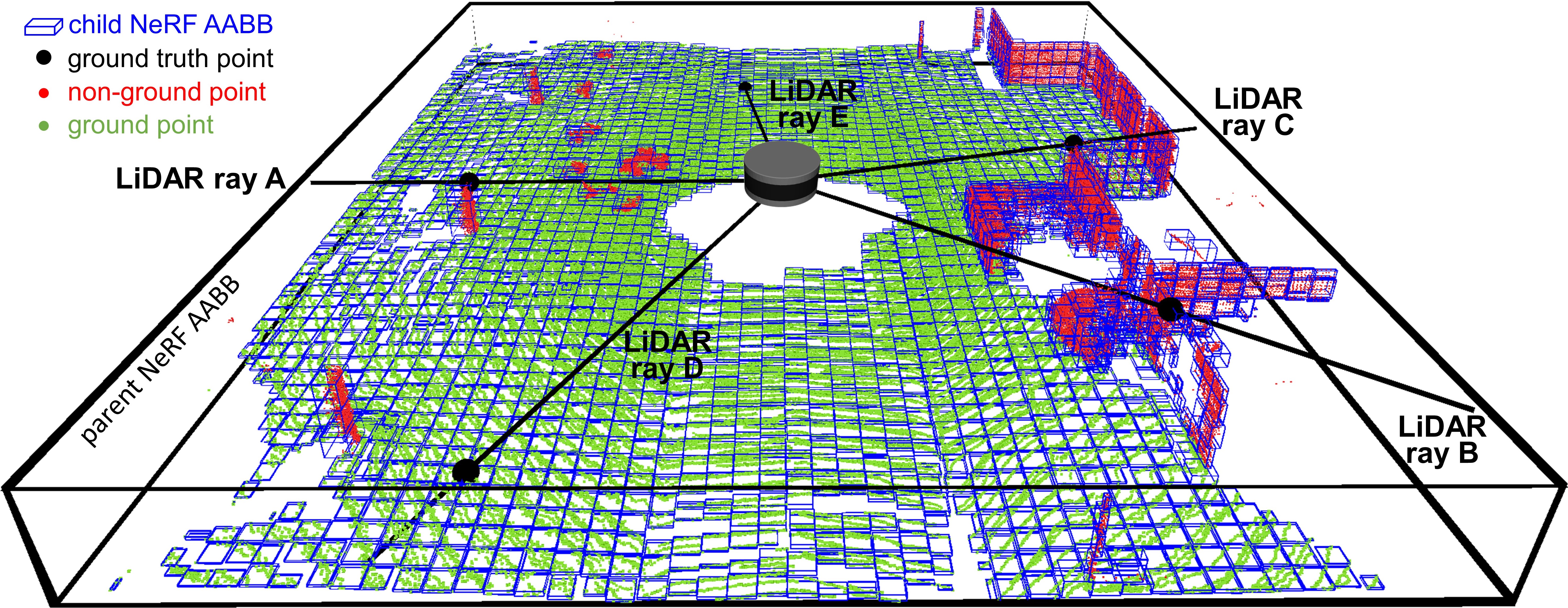}%
		\label{fig_infer_1}}
	\hspace{0.01in}		
	\subfigure[Weight distribution and depth inference along 5 LiDAR rays.]{\includegraphics[width=3.2in]{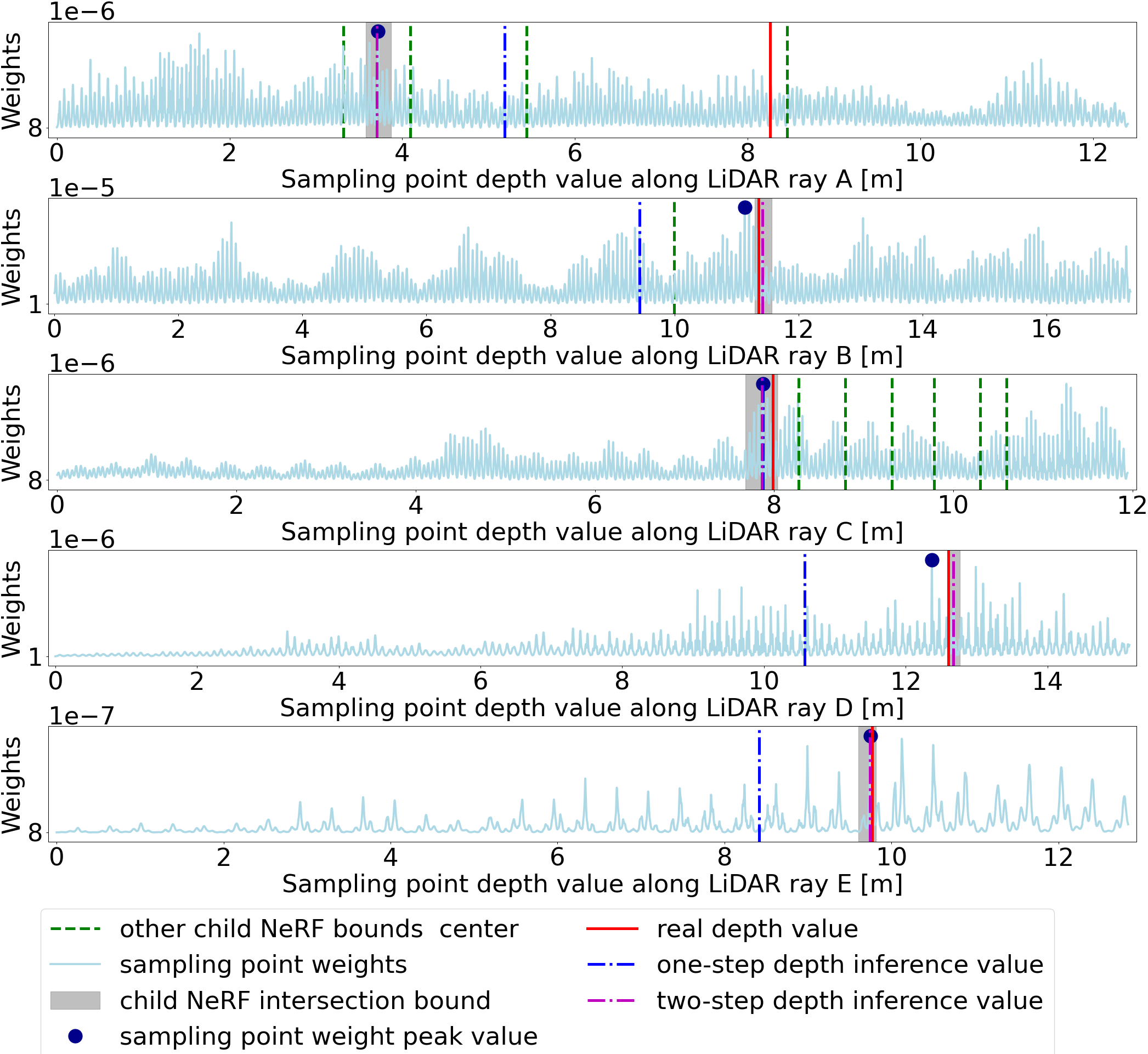}%
		\label{fig_infer_2}}
	\hfil	
	\caption{Parent-child NeRF's two-step depth inference effect illustration. The five subfigures in Fig.~\ref{fig_infer}(b) represent depth value inference results for the five LiDAR rays in Fig.~\ref{fig_infer}(a), where the weight distribution data comes from our proposed PC-NeRF model trained on the KITTI 00 sequence 1151-1200 frame scene in Sec.~\ref{sec:Evaluating}.}
	\label{fig_infer}
\end{figure}

Most current depth inference methods are one-step depth inference methods, similar to Eq.~\ref{equation7_2}. In contrast, we provide a two-step depth inference method (as Fig.~\ref{framework}(c) illustrates) to infer more accurately, especially training with sparse LiDAR frames.
We first search in the parent NeRF AABB to acquire the child NeRF AABBs potentially intersecting the LiDAR ray $\mathbf{r}_{ij}$ and then conduct further inference in the child NeRF AABB's near and far bounds $[\hat{n_{ij}^{c}}, \hat{f_{ij}^{c}}]$ intersecting with the LiDAR ray, as shown in Fig.~\ref{fig_infer}(a). To this end, we first select the child NeRF whose AABB outer sphere intersects the ray and then use the Axis Aligned Bounding Box intersection test proposed by Haines \cite{haines1989essential}, which can readily process millions of voxels or AABBs in real time \cite{liu2020neural}. If the LiDAR ray does not intersect any child NeRF AABB, the space extent of all child NeRF AABBs for this LiDAR ray should be slightly inflated in incremental steps, and the above inference should be performed again.

To mitigate the risk of misinterpreting free space as object space, we constrain the inferred depth values within the near and far bounds of a single child NeRF, as depicted in the first subfigure of Fig.~\ref{fig_infer}(b).
In complex scenes with LiDAR rays intersecting multiple child NeRF AABBs, we select the AABB containing the peak weight value (calculated by Eq.~\ref{equation3}). This choice is illustrated in the first and third subfigures of Fig.~\ref{fig_infer}(b), considering the LiDAR ray's inability to penetrate opaque objects during emission.
Besides, when the peak weight value is not in any child NeRF AABB, we opt for the child NeRF with the maximum weight integration $W$ (from Eq.~\ref{equation9_1}), i.e., the most probable existence range of an object on the LiDAR ray, as shown in the second subfigure of Fig.~\ref{fig_infer}(b). 
When the LiDAR ray only intersects one child NeRF AABB, we can directly select it as the object area, as shown in the last two subfigures of Fig.~\ref{fig_infer}(b). 
Note that if the child NeRF weight integration \(W\) is minimal, the LiDAR ray doesn't intersect any child NeRF, and no depth value needs to be inferred.
The inferred depth value of each LiDAR ray is calculated by Eq.~\ref{equation10}.
\begin{equation}
	\label{equation9_1}
	W = \int_{\hat{n_{ij}^{c}}}^{\hat{f_{ij}^{c}}} w(t)dt
\end{equation}
\begin{equation}
	\label{equation10}
	\hat{d_{ij}} =\frac{{\int_{\hat{n_{ij}^{c}}}^{\hat{f_{ij}^{c}}}w(t)\cdot tdt}}
	{{\int_{\hat{n_{ij}^{c}}}^{\hat{f_{ij}^{c}}} w(t)dt}}
\end{equation}

\section{Experiments}
\subsection{Experiment Setups}\label{sec:setting}
\textbf{Datasets}: We evaluate our proposed PC-NeRF using 13 data sequences from two publicly available outdoor datasets, including the MaiCity \cite{vizzo2021poisson} and KITTI odometry datasets \cite{geiger2012we}. The 00 and 01 sequences of the MaiCity dataset contain 64-beam noise-free synthetic LiDAR frames in virtual urban-like environments. The KITTI odometry dataset also contains 64-beam LiDAR data collected by vehicles driving in real-world environments and provides a localization benchmark with ground-truth vehicle poses. The semantic labels of point clouds in KITTI are from SemanticKITTI \cite{behley2019semantickitti}, which further filters out movable objects. This filtering enhances the accuracy and stability of 3D reconstruction and novel view synthesis, simplifies processing, and improves computational efficiency. 

\textbf{Frame sparsity}:
Frame sparsity represents the proportion of the test set (unavailable during training) when dividing the LiDAR dataset into training and test sets. 
Increased frame sparsity implies fewer LiDAR frames for training and more for model testing, posing heightened challenges across various tasks. Following the dataset splitting method employed in previous works like NeRF-LOAM \cite{Deng_2023_ICCV} and NFL \cite{Huang2023nfl}, we initially take one test frame from every five frames, setting the frame sparsity to 20\,\%. 
Following that, we systematically vary frame sparsity by selecting test frames as follows: one from every four frames (25\,\%), one from every three frames (33\,\%), one from every two frames (50\,\%), two from every three frames (67\,\%), three from every four frames (75\,\%), and four from every five frames (80\,\%).
Therefore, the test frame selection for $\left \{20\%, 25\%, 33\%, 50\%\right \}$ frame sparsities shows a regular increase, as does the test frame selection for $\left \{50\%, 67\%, 75\%, 80\%\right \}$ frame sparsities.
In addition, it is reasonable to use 50\,\% (the frame sparsity threshold, i.e., the same number of LiDAR frames for model training and model performance testing) as a transition from $\left \{20\,\%, 25\,\%, 33\,\%\right \}$  to $\left \{67\,\%, 75\,\%, 80\,\%\right \}$. We further explore extreme dataset splitting by selecting nine test frames from every ten frames, resulting in a frame sparsity of 90\,\%. In summary, selecting these eight frame sparsities of $\left \{20\,\%, 25\,\%, 33\,\%, 50\,\%, 67\,\%, 75\,\%, 80\,\%, 90\,\%\right \}$ is meaningful and represents the range of 0\,\% to 100\,\% well.

\textbf{Metrics}: 
We evaluate our method's performance in novel LiDAR view synthesis, single-frame 3D reconstruction, and 3D scene reconstruction.
We use the error metrics presented in IR-MCL \cite{kuang2022ir} and UrbanNeRF \cite{rematas2022urban} to evaluate novel LiDAR view synthesis and single-frame 3D reconstruction performance. 
By comparing each synthesized LiDAR frame with its corresponding real LiDAR frame on the test set and averaging the metrics across all test frames, we report the average absolute error of LiDAR ray depth (Dep. Err.\,[$\mathrm{m}$]), the average accuracy of LiDAR ray depth at 0.2\,$\mathrm{m}$ threshold (Dep. Acc@0.2$\mathrm{m}$\,[\%]), chamfer distance (CD\,[$\mathrm{m}$]), and F-score at 0.2\,$\mathrm{m}$ threshold (F@0.2$\mathrm{m}$).
For 3D scene reconstruction, we concatenate individual synthesized and real LiDAR frames using the test set poses to generate the reconstructed and real LiDAR point cloud maps.
We use the reconstruction metrics commonly used in most reconstruction methods \cite{Deng_2023_ICCV, zhong2023shine, mescheder2019occupancy, vizzo2021poisson}, i.e., accuracy (Acc.\,[$\mathrm{m}$]), completion (Comp.\,[$\mathrm{m}$]), chamfer distance (Map CD\,[$\mathrm{m}$]), and F-score at 0.2\,$\mathrm{m}$ threshold (Map F@0.2$\mathrm{m}$), to evaluate the 3D scene reconstruction results between the reconstructed and real LiDAR point cloud maps. Given the inherent interdependence between 3D scene reconstruction and single-frame 3D reconstruction, the experiments in Sec.~\ref{sec:Evaluating}, Sec.~\ref{sec:Lesser point clouds for 3D reconstruction}, and Sec.~\ref{sec:Ablation} focus primarily on evaluating metrics related to single-frame 3D Reconstruction.

\textbf{Baselines}: A standard pipeline for generating new LiDAR views is constructing a 3D point cloud map and then using the ray-casting approach \cite{thrun2002probabilistic} to query new point clouds from the map. We implement this pipeline, which voxelizes the 3D point cloud map (voxel size: 0.05\,$\mathrm{m}$) into a 3D voxel map to speed up the query but may slightly reduce accuracy, as a baseline method named \textbf{MapRayCasting}. In addition, we also extend the original NeRF model proposed by Mildenhall \cite{mildenhall2021nerf} by replacing a camera ray with a LiDAR ray as IR-MCL \cite{kuang2022ir}, named \textbf{OriginalNeRF}. For the memory consumption, OriginalNeRF and PC-NeRF include the model file and the child NeRFs bounds file, while MapRayCasting includes a voxel map. Furthermore, we also employ ground truth poses in \textbf{NeRF-LOAM} \cite{Deng_2023_ICCV} to reconstruct the mesh map of the environments. In the whole execution of NeRF-LOAM, a shared network of just two fully connected layers with 256 units per layer is used. Therefore, it is difficult for NeRF-LOAM to perform the novel view synthesis task using the network weights obtained from training. Accordingly, we use the mesh obtained from NeRF-LOAM for the 3D scene reconstruction task. 

\textbf{Training details}: We train our proposed PC-NeRF and all the baselines with an NVIDIA GeForce RTX 3090 and an Intel i9-11900K CPU and use Adam \cite{kingma2014adam} as the training optimizer. Since reconstructing and storing large-scale outdoor scenes as NeRF models require a relatively lengthy training period \cite{martin2021nerf, rematas2022urban, turki2022mega, tancik2022block}, we aim to achieve optimal results with minimal training epochs, particularly emphasizing one epoch for swift deployment.
After training with the current number of epochs, if invalid inferences occur on certain LiDAR rays during the two-step depth inference method, we will increment the number of epochs, e.g., 2, 5, or 10. 
Besides, considering the relatively time-consuming nature of each epoch's training, as indicated in Tab.~\ref{tab:Parent-child NeRF Inference Effect} and~\ref{tab:Lesser}, we have chosen not to include experimental results for the maximum possible epochs in the paper.
Based on extensive experimental testing, the initial learning rate is set to $4\times 10^{-5}$ and adjusted using Pytorch's MultiStepLR strategy with milestones at $[5,120]$ and an adjustment factor of $0.1$. The experimental results on the KITTI and MaiCity datasets, including Sec.~\ref{sec:Evaluating}, Sec.~\ref{sec:Lesser point clouds for 3D reconstruction}, and Sec.~\ref{sec:Ablation}, demonstrate that our proposed PC-NeRF achieves satisfactory results with only one training epoch in most scenes. In Sec.~\ref{sec:Evaluating} and Sec.~\ref{sec:Lesser point clouds for 3D reconstruction}, $\lambda_{\mathrm{pd}}$, $\lambda_{\mathrm{cf}}$, $\lambda_{\mathrm{cd}}$, $\lambda_{\mathrm{in}}$, and $\gamma$ of our proposed PC-NeRF are set to $1$, $10^{6}$, $10^{5}$, $0.1$, and $2.0\,m$, respectively. This group of parameters is not specifically tuned for a single scene but is valid for all experimental scenes, and the corresponding ablation studies are shown in Sec.~\ref{sec:Ablation}. Both our PC-NeRF and OriginalNeRF use the hierarchical volume sampling strategy along the LiDAR ray proposed in NeRF \cite{mildenhall2021nerf}, where the points number $N_c$ and $N_f$ for coarse and fine sampling along the ray are 768 and 1536, respectively. However, for coarse sampling, our PC-NeRF uses the Child NeRF segmented sampling proposed in Sec.~\ref{sec:LiDAR loss}, while OriginalNeRF samples uniformly along the LiDAR rays.

\subsection{Evaluation for Novel LiDAR View Synthesis and 3D Reconstruction in Different Scales}\label{sec:Evaluating}
\begin{figure*}[!t]
\centering
\subfigure[Input LiDAR frames: used as the training set for model training (OriginalNeRF and PC-NeRF) or for constructing 3D voxel maps (MapRayCasting).]{\includegraphics[width=7.0in]{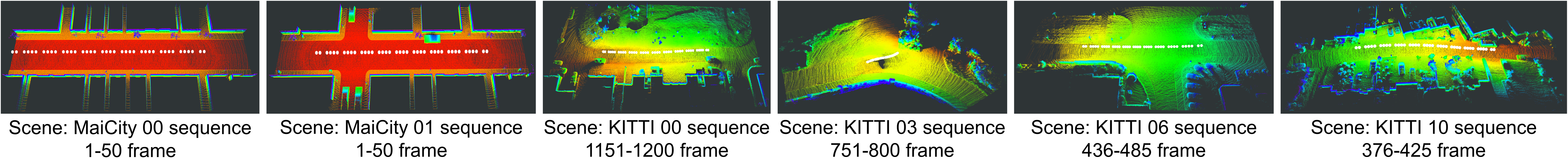}%
	\label{fig_cd_F_score2_tmp}}		
\hspace{0.01in}		
\subfigure[3D scene reconstruction performance of PC-NeRF, OriginalNeRF, and MapRayCasting.]{\includegraphics[width=7.0in]{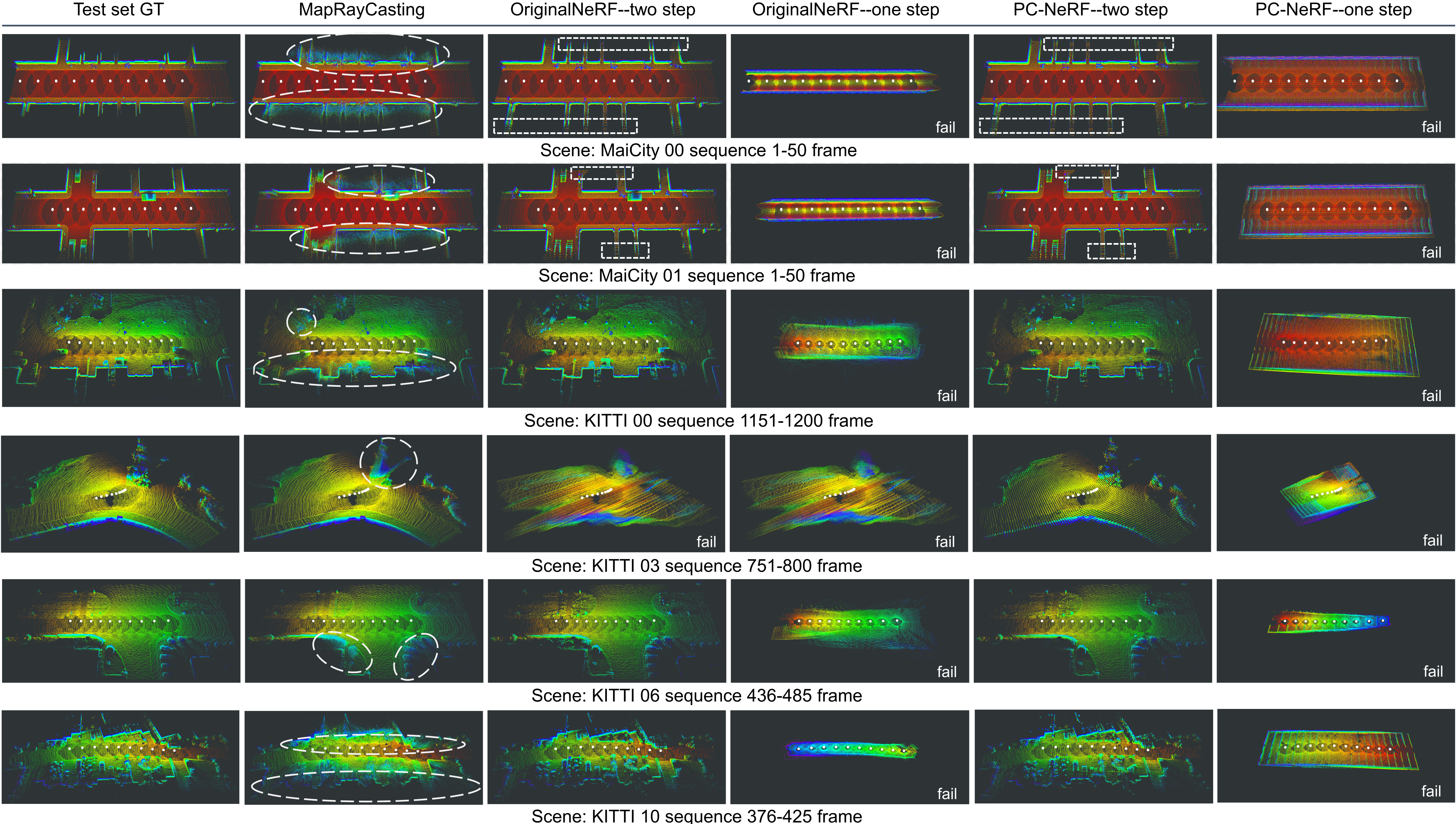}%
	\label{fig_source_merge1_tmp}}
\caption{3D scene reconstruction on the MaiCity and KITTI datasets. Each subfigure represents the result of concatenating multiple LiDAR frames using real poses. The white dots in each subfigure represent the LiDAR positions of each frame. In Fig. (b): ``one/two-step" denotes the one/two-step depth inference method. Fig. (b) illustrates the inference results corresponding to Tab.~\ref{tab:Parent-child NeRF Inference Effect}.  The ``PC-NeRF--two step" column in Fig. (b) corresponds to the rows with 20\,\% frame sparsity in Tab.~\ref{tab:Lesser}.}
\label{Inference-effects}
\end{figure*}

We qualitatively and quantitatively evaluate the inference accuracy and deployment potential of our PC-NeRF across different scales.
For small-scale evaluation, we use 50 consecutive LiDAR frames from the MaiCity and KITTI datasets as a single scene, training and evaluating each model on this scene. The frame sparsity is set to 20\,\% and the experimental results are presented in Fig.~\ref{Inference-effects}, Tab.~\ref{tab:Parent-child NeRF Inference Effect}, and Tab.~\ref{tab2}.

As shown by the white dashed ellipses in the second column of Fig.~\ref{Inference-effects}(b), MapRayCasting successfully reconstructs the overall environment but introduces shadow artifacts. 
In the third to sixth column of Fig.~\ref{Inference-effects}(b), the one-step depth inference method fails to reconstruct the environment, while the two-step depth inference method outlined in Sec.~\ref{sec:Two-step Range Value Inference} exhibits effective performance.
Besides, employing the two-step depth inference method, our proposed PC-NeRF and OriginalNeRF additionally infer structures that do not belong to the test frames (as shown by the white dashed box in the third and fifth columns of Fig.~\ref{Inference-effects}(b)) but are genuinely present in the actual scene (as shown in Fig.~\ref{Inference-effects}(a)) which is more beneficial for 3D scene reconstruction. 
As demonstrated in the third column of the fourth row in Fig.~\ref{Inference-effects}(b), OriginalNeRF faces challenges in scene reconstruction with the two-step depth inference. 
Additionally, due to the invalid inferences on some LiDAR rays when utilizing the two-step depth inference method, we rely on the remaining valid inferences for quantitative evaluation in Tab.~\ref{tab:Parent-child NeRF Inference Effect}. 	
These challenges and invalid inferences occur because OriginalNeRF cannot quickly obtain the approximate environment distribution during training, leading to the failure of the first step in the two-step depth inference method. 
To sum up, implementing the two-step depth inference method with PC-NeRF yields the best qualitative results.

As shown in Tab.~\ref{tab:Parent-child NeRF Inference Effect}, 
our proposed PC-NeRF outperforms MapRayCasting and OriginalNeRF, demonstrating superior accuracy in novel LiDAR view synthesis and single-frame 3D reconstruction.
Besides, our proposed PC-NeRF demonstrates excellent deployment potential as it yields superior results with just one epoch of training, maintaining consistent training set across various scenes. 
Tab.~\ref{tab:Parent-child NeRF Inference Effect} shows that by combining training time, memory consumption, novel LiDAR view synthesis accuracy, and single-frame 3D reconstruction accuracy, our proposed PC-NeRF is far superior to OriginalNeRF and MapRayCasting. 
Additionally, in Tab.~\ref{tab2} (rows with a frame sparsity of 20\,\%), we present the 3D scene reconstruction results for the experimental groups (``PC-NeRF" + ``two-step") from Tab.~\ref{tab:Parent-child NeRF Inference Effect}, showcasing the superior performance of our proposed PC-NeRF over NeRF-LOAM in 3D scene reconstruction accuracy on the KITTI and MaiCity datasets.
This superior performance is because the network structure of NeRF-LOAM is lightweight, prioritizing the real-time implementation of the algorithm while slightly reducing the 3D scene reconstruction accuracy.
In summary, applying the two-step depth inference method to PC-NeRF yields the highest quantitative results.

\begin{figure}[!t]
\centering
\subfigure[Real LiDAR point clouds of all test frames.]{\includegraphics[width=3.5in]{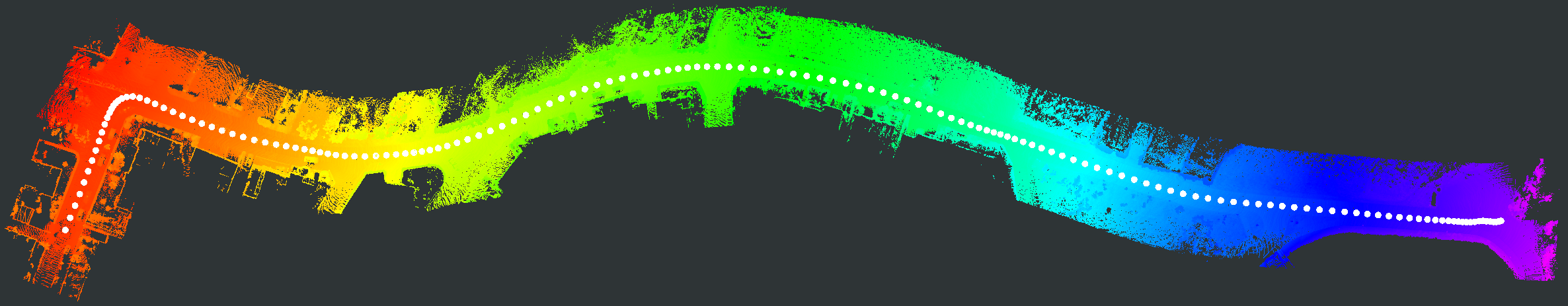}%
	\label{fig_source_merge}}
\hspace{0.01in}		
\subfigure[Reconstructed LiDAR point clouds by our proposed PC-NeRF.]{\includegraphics[width=3.5in]{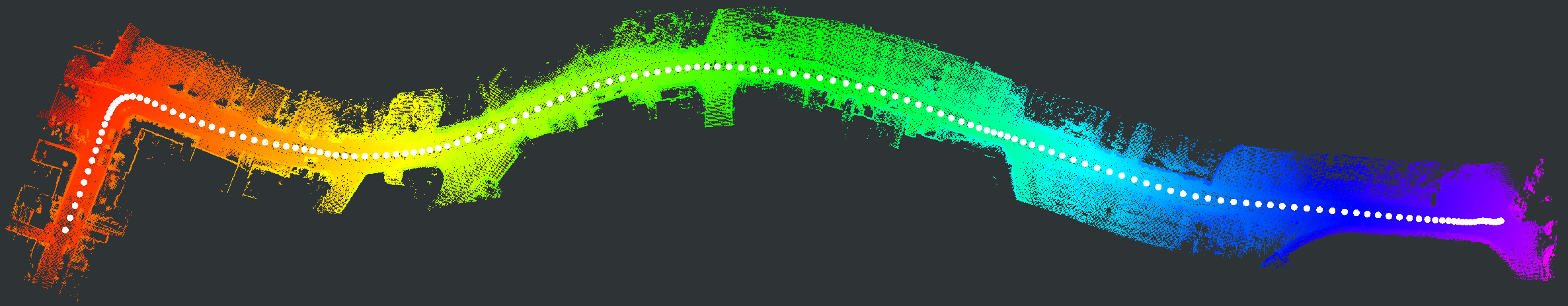}	
	\label{fig_prediction_pc_nerf_merge}}
\hspace{0.01in}		
\subfigure[Single-frame 3D reconstruction accuracy of our proposed PC-NeRF.]{\includegraphics[width=3.5in]{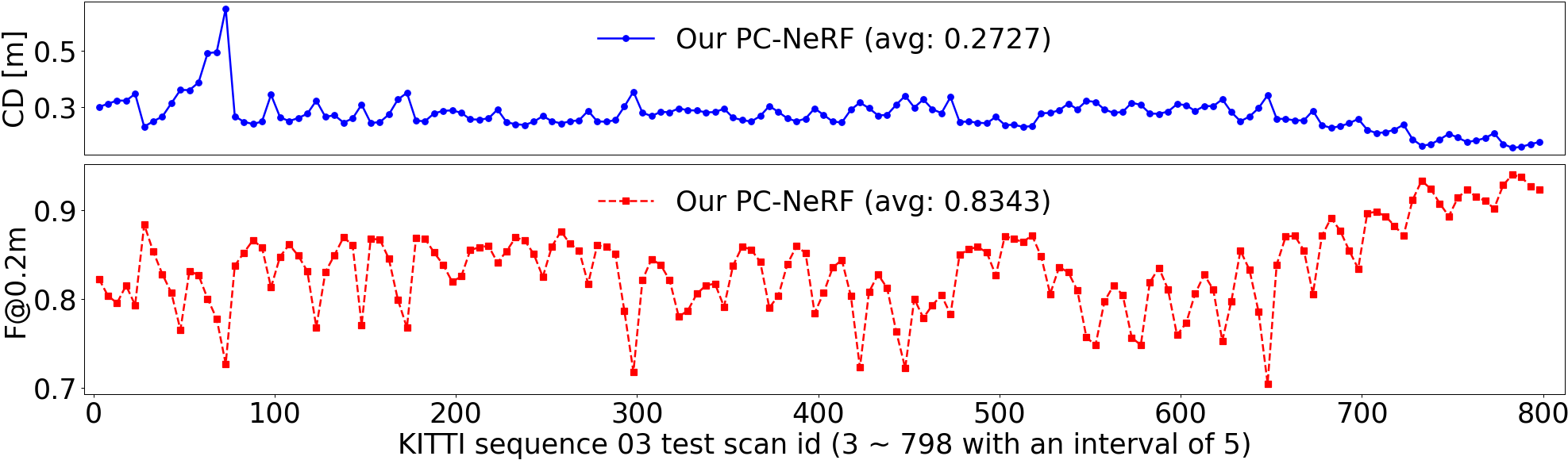}%
	\label{fig_cd_F_score2}}
\hspace{0.01in}			
\caption{3D reconstruction effect using our proposed PC-NeRF on KITTI 03 sequence. The white dots in Fig.~\ref{fig_motivation}(a) and Fig.~\ref{fig_motivation}(b) indicate the LiDAR position of each frame, corresponding to each data point in Fig.~\ref{fig_motivation}(c). As qualitative and quantitative results are shown in the three sub-figures, our proposed PC-NeRF has high 3D reconstruction accuracy in the KITTI 03 sequence.}
\label{fig_motivation}
\end{figure}

For large-scale evaluation, according to Sec.~\ref{sec:parent NeRF}, we divide the KITTI 03 sequence (800 consecutive frames, $555\times 120\times 7.2\,{\mathrm{m}}^{3}$) into 32 sequential blocks. The spatial extent of each block is about $61.5\times 42.5\times 3\,{\mathrm{m}}^{3}$, overlapping with that of neighboring blocks. In the 30th block, our proposed PC-NeRF trains three epochs because one or two epochs of training result in low inference accuracy with two-step depth inference. Additionally, our proposed PC-NeRF trains only one epoch on all the remaining 31 blocks. As a comparison, OriginalNeRF trains three epochs on all 32 blocks. 
It is mentioned in the previous two paragraphs that the two-step depth inference method for OriginalNeRF may result in invalid inferences for some LiDAR rays, whereas for PC-NeRF, it performs well.
Therefore, in Tab.~\ref{tab:table-KITTI-sequence03}, we use the one-step depth inference method for OriginalNeRF and apply the two-step depth inference method for PC-NeRF.
From Tab.\ref{tab:table-KITTI-sequence03} and Fig.\ref{fig_motivation}, it is evident that our proposed PC-NeRF is trained to achieve high accuracy in novel LiDAR view synthesis and single-frame 3D reconstruction. This further underscores its promising potential for deployment in large-scale environments.

\newcommand{\uparrowvalueb}[1]{\makebox[0pt][l]{#1}\hspace{0.5cm}\rotatebox[origin=c]{0}{$\uparrow$}}
\newcommand{\uparrowvaluec}[1]{\makebox[0pt][l]{#1}\hspace{1.3cm}\rotatebox[origin=c]{0}{$\uparrow$}}
\newcommand{\downarrowvaluec}[1]{\makebox[0pt][l]{#1}\hspace{0.9cm}\rotatebox[origin=c]{180}{$\uparrow$}}
\newcommand{\downarrowvalued}[1]{\makebox[0pt][l]{#1}\hspace{0.4cm}\rotatebox[origin=c]{180}{$\uparrow$}}
\begin{table*}
\centering
\renewcommand\arraystretch{1.5}		
\caption{Novel LiDAR view synthesis and single-frame 3D reconstruction on small-scale scenes (frame sparsity: 20\,$\%$, bolded: best results for the corresponding scene/dataset)}
{\fontsize{7.35}{8}\selectfont 		
	\begin{tabular}{ccccccccccc}
		\cline{1-11}	
		\midrule		
		\multicolumn{1}{c}{\begin{tabular}[c]{@{}c@{}}Dataset\end{tabular}} & \multicolumn{2}{c}{\begin{tabular}[c]{@{}c@{}}Method\end{tabular}} & 
		\multicolumn{1}{c}{\begin{tabular}[c]{@{}c@{}}Train\\epoch\end{tabular}} & 
		\multicolumn{1}{c}{\begin{tabular}[c]{@{}c@{}}Train time/\\epoch\,$[$min$]$\end{tabular}} &			
		\multicolumn{1}{c}{\begin{tabular}[c]{@{}c@{}}Memory\\consumption\end{tabular}} & 
		\multicolumn{1}{c}{\begin{tabular}[c]{@{}c@{}}Inference\\method\end{tabular}}& 
		\multicolumn{1}{c}{\downarrowvaluec{\begin{tabular}[c]{@{}c@{}}Dep.\\Err.\,[$\mathrm{m}$]\end{tabular}}}&
		\multicolumn{1}{c}{\uparrowvaluec{\begin{tabular}[c]{@{}c@{}}Dep. Acc@\\0.2m\,[\%]\end{tabular}}}&		
		\multicolumn{1}{c}{\downarrowvalued{\begin{tabular}[c]{@{}c@{}}CD \\$[$m$]$\end{tabular}}}	& 
		\multicolumn{1}{c}{\uparrowvalueb{\begin{tabular}[c]{@{}c@{}}F@\\0.2m\end{tabular}}}	 \\ 
		\cline{1-11}
		\midrule	
		\multicolumn{1}{c}{\multirow{5}[0]{*}{\begin{tabular}[c]{@{}c@{}}MaiCity\\00-01\\sequence\end{tabular}}} & \multicolumn{2}{c}{MapRayCasting} &   \multicolumn{1}{c}{-}    &    \multicolumn{1}{c}{-}   & \multicolumn{1}{c}{5.8\,MB} &    \multicolumn{1}{c}{-}   & 0.390  & 81.760  & 0.261  & 0.863  \\
		\cline{2-11}	
		& \multicolumn{2}{c}{\multirow{2}[0]{*}{OriginalNeRF}} & \multirow{2}[0]{*}{1} & \multirow{2}[0]{*}{71} & \multicolumn{1}{c}{12.1\,MB} & one-step & 3.581  & 0.154  & 3.336  & 0.000  \\
		& \multicolumn{2}{c}{} &     &      & 12.1\,MB\,+\,300.7\,KB & {two-step} & 0.505  & 85.008  & 0.265  & 0.921  \\
		\cline{2-11}	
		& \multicolumn{2}{c}{\multirow{2}[0]{*}{\textbf{PC-NeRF}}} & \multirow{2}[0]{*}{1}      & \multirow{2}[0]{*}{92} & \multicolumn{1}{c}{12.1\,MB} & one-step & 1.890  & 2.382  & 1.367  & 0.063  \\
		& \multicolumn{2}{c}{} &       &       & 12.1\,MB\,+\,300.7\,KB & \textbf{two-step} & \textbf{0.347} & \textbf{87.877} & \textbf{0.179} & \textbf{0.945} \\
		\cline{1-11}
		\multicolumn{1}{c}{\multirow{5}[0]{*}{\begin{tabular}[c]{@{}c@{}}KITTI\\00-10\\sequence\end{tabular}}} & \multicolumn{2}{c}{MapRayCasting} &   \multicolumn{1}{c}{-}    &   \multicolumn{1}{c}{-}    & \multicolumn{1}{c}{17.5\,MB} &   \multicolumn{1}{c}{-}    & \textbf{0.461} & \textbf{68.754} & 0.265  & 0.837  \\
		\cline{2-11}	
		& \multicolumn{2}{c}{\multirow{2}[0]{*}{OriginalNeRF}} & \multirow{2}[0]{*}{1} & \multirow{2}[0]{*}{58} & \multicolumn{1}{c}{12.1\,MB} & one-step & 4.111  & 1.944  & 2.803  & 0.078  \\
		& \multicolumn{2}{c}{} &       &       & 12.1\,MB\,+\,917.6\,KB & {two-step} & 0.749  & 52.658  & 0.297  & 0.818  \\
		\cline{2-11}	
		& \multicolumn{2}{c}{\multirow{2}[0]{*}{\textbf{PC-NeRF}}} &  \multirow{2}[0]{*}{1}     & \multirow{2}[0]{*}{76} & \multicolumn{1}{c}{12.1\,MB} & one-step & 5.366  & 1.281  & 4.478  & 0.039  \\
		& \multicolumn{2}{c}{} &       &       & 12.1\,MB\,+\,917.6\,KB & \textbf{two-step} & 0.592  & 59.149  & \textbf{0.244} & \textbf{0.863} \\
		\cline{1-11}	
		\midrule			
	\end{tabular}%
}
\label{tab:Parent-child NeRF Inference Effect}%
\end{table*}%

\newcommand{\uparrowvaluea}[1]{\makebox[0pt][l]{#1}\hspace{0.5cm}\rotatebox[origin=c]{0}{$\uparrow$}}
\newcommand{\downarrowvaluea}[1]{\makebox[0pt][l]{#1}\hspace{0.5cm}\rotatebox[origin=c]{180}{$\uparrow$}}
\newcommand{\downarrowvalueb}[1]{\makebox[0pt][l]{#1}\hspace{0.65cm}\rotatebox[origin=c]{180}{$\uparrow$}}
\begin{table}[htbp]
\renewcommand\arraystretch{1.4}		
\centering
\caption{3D scene reconstruction on the KITTI and MaiCity dataset (type of depth inference for PC-NeRF: two-step)}
{\fontsize{7}{8}\selectfont 	
	\begin{tabular}{ccccccc}
		\cline{1-7}	 
		\midrule		
		\multicolumn{1}{c}{\begin{tabular}[c]{@{}c@{}}Dataset\end{tabular}} &
		\multicolumn{1}{c}{Method} 		& 		 \multicolumn{1}{c}{\begin{tabular}[c]{@{}c@{}}Frame\\sparsity\\$[\%]$\end{tabular}}&  
		\multicolumn{1}{c}{\downarrowvaluea{\begin{tabular}[c]{@{}c@{}}Acc.\\$[$m$]$\end{tabular}}} &		
		\multicolumn{1}{c}{\downarrowvalueb{\begin{tabular}[c]{@{}c@{}}Comp.\\$[$m$]$\end{tabular}}} &				
		\multicolumn{1}{c}{\downarrowvaluea{\begin{tabular}[c]{@{}c@{}}Map\\CD\\$[$m$]$\end{tabular}}}	& 
		\multicolumn{1}{c}{\uparrowvaluea{\begin{tabular}[c]{@{}c@{}}Map\\F@\\0.2m\end{tabular}}}	 \\
		\cline{1-7}	 
		\midrule		
		\multicolumn{1}{c}{\multirow{6}[0]{*}{\begin{tabular}[c]{@{}c@{}}MaiCity\\00-01\\sequence\end{tabular}}}& \multicolumn{1}{c}{\multirow{3}[0]{*}{\begin{tabular}[c]{@{}c@{}}NeRF-\\LOAM\end{tabular}}} & 20    & 0.029  & 0.032  & 0.030  & 99.253  \\
		&       & 67    & 0.026  & 0.066  & 0.046  & 98.045  \\
		&       & 90    & 0.027  & 0.122  & 0.074  & 95.777  \\
		\cline{2-7}	 
		& \multicolumn{1}{c}{\multirow{3}[0]{*}{\begin{tabular}[c]{@{}c@{}}PC-\\NeRF\end{tabular}}} & 20    & 0.023  & 0.028  & 0.025  & 99.836  \\
		&       & 67    & 0.023  & 0.017  & 0.020  & 99.883  \\
		&       & 90    & 0.024  & 0.025  & 0.024  & 99.500  \\
		\cline{1-7}	 
		\multicolumn{1}{c}{\multirow{6}[0]{*}{\begin{tabular}[c]{@{}c@{}}KITTI\\00-10\\sequence\end{tabular}}} & \multicolumn{1}{c}{\multirow{3}[0]{*}{\begin{tabular}[c]{@{}c@{}}NeRF-\\LOAM\end{tabular}}} & 20    & 0.062  & 0.174  & 0.118  & 86.477  \\
		&       & 67    & 0.060  & 0.182  & 0.121  & 86.253  \\
		&       & 90    & 0.060  & 0.265  & 0.162  & 79.661  \\
		\cline{2-7}	 		
		& \multicolumn{1}{c}{\multirow{3}[0]{*}{\begin{tabular}[c]{@{}c@{}}PC-\\NeRF\end{tabular}}} & 20    & 0.037  & 0.090  & 0.063  & 96.141  \\
		&       & 67    & 0.034  & 0.070  & 0.052  & 97.215  \\
		&       & 90    & 0.032  & 0.069  & 0.051  & 96.868  \\
		\cline{1-7}	 
		\midrule		
	\end{tabular}
}
\label{tab2}
\end{table}

\newcommand{\downarrowvalueg}[1]{\makebox[0pt][l]{#1}\hspace{0.8cm}\rotatebox[origin=c]{180}{$\uparrow$}}
\newcommand{\uparrowvalueh}[1]{\makebox[0pt][l]{#1}\hspace{1.15cm}\rotatebox[origin=c]{0}{$\uparrow$}}
\begin{table}[!t]
\renewcommand\arraystretch{1.4}	
\caption{novel LiDAR view synthesis and single-frame 3D reconstruction on large-scale scenes (frame sparsity: 20\,$\%$)}		
{\fontsize{7}{8}\selectfont 
	\begin{tabularx}{\columnwidth}{>{\centering\arraybackslash}p{0.05in}>{\centering\arraybackslash}p{0.05in}>{\centering\arraybackslash}p{0.05in}>{\centering\arraybackslash}p{0.05in}>{\centering\arraybackslash}p{0.05in}>{\centering\arraybackslash}p{0.1in}}
		\cline{1-6}	
		\multicolumn{1}{c}{Method}        
		&\multicolumn{1}{c}{\begin{tabular}[c]{@{}c@{}}Depth\\inference\end{tabular}}&
		\multicolumn{1}{c}{\downarrowvalueg{\begin{tabular}[c]{@{}c@{}}Dep.\\Err.\,[$\mathrm{m}$]\end{tabular}}}&
		\multicolumn{1}{c}{\uparrowvalueh{\begin{tabular}[c]{@{}c@{}}Dep. Acc@\\0.2m\,[\%]\end{tabular}}}&
		\multicolumn{1}{c}{\downarrowvalued{\begin{tabular}[c]{@{}c@{}}CD\\$[$m$]$\end{tabular}}} & \multicolumn{1}{c}{\uparrowvalueb{\begin{tabular}[c]{@{}c@{}}F@\\0.2m\end{tabular}}} \\	\cline{1-6}		
		\multicolumn{1}{c}{\begin{tabular}[c]{@{}c@{}}MapRayCasting\end{tabular}}
		&\multicolumn{1}{c}{-}
		&\multicolumn{1}{c}{0.898}  &\multicolumn{1}{c}{34.812}  &\multicolumn{1}{c}{0.452}  &\multicolumn{1}{c}{0.729} \\
		\multicolumn{1}{c}{OriginalNeRF} & \multicolumn{1}{c}{one-step}  &\multicolumn{1}{c}{2.441} &\multicolumn{1}{c}{8.290} &\multicolumn{1}{c}{1.438}  &\multicolumn{1}{c}{0.271} \\
		\multicolumn{1}{c}{\textbf{PC-NeRF}} & \multicolumn{1}{c}{two-step} &\multicolumn{1}{c}{\textbf{0.658}} &\multicolumn{1}{c}{\textbf{43.556}} &   \multicolumn{1}{c}{\textbf{0.273}} &\multicolumn{1}{c}{\textbf{0.834}}  \\   \cline{1-6}	                             
	\end{tabularx}
}
\label{tab:table-KITTI-sequence03}%
\end{table}

\subsection{Evaluation for Novel LiDAR View Synthesis and 3D Reconstruction using Sparse LiDAR Frames}\label{sec:Lesser point clouds for 3D reconstruction}
\begin{figure*}[!t]
\centering
\includegraphics[width=7.0in]{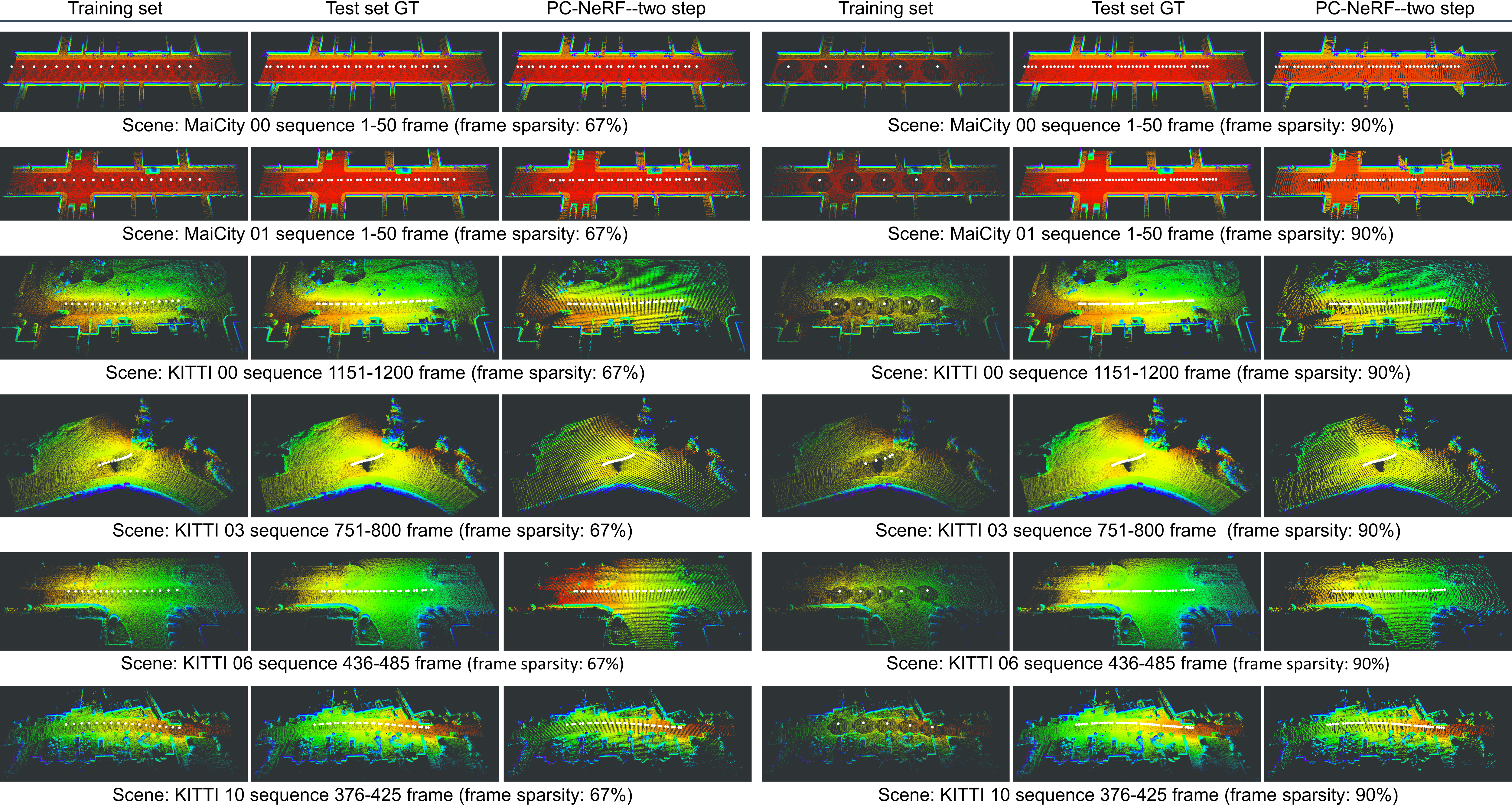}%
\caption{3D scene reconstruction using sparse LiDAR frames on the MaiCity and KITTI datasets. ``two-step" denotes the two-step depth inference method. Each subfigure represents the result of concatenating multiple LiDAR frames using real poses. The white dots in each subfigure represent the LiDAR positions of each frame.}
\label{testsparse_1}
\end{figure*}

To assess the novel LiDAR view synthesis and 3D reconstruction capabilities of our proposed PC-NeRF using sparse LiDAR frames, we perform experiments on the MaiCity and KITTI datasets, treating 50 consecutive frames as a scene. Each PC-NeRF model is trained and evaluated based on a single scene.
The sparse LiDAR frames are obtained by adjusting the frame sparsity, as described in the second paragraph of Sec.~\ref{sec:setting}. 
As seen in Tab.~\ref{tab:Lesser}, Fig.~\ref{fig:fig_motivation2}, and Fig.~\ref{testsparse_1}, with only one epoch training, our proposed PC-NeRF can reconstruct environments very well, even if the frame sparsity reaches 67\,$\%$. 
In Tab.~\ref{tab:Lesser}, specifically from rows 2 to 25, it is evident that as the proportion of the training set in the entire dataset gradually decreases from 4/5 to 1/10, indicating an increase in training difficulty, the three scenes start requiring more epochs at 75\,\% ($>$\,67\,\%), 67\,\%, and 67\,\% of frame sparsity, respectively, to achieve valid inference by the two-step depth inference over all LiDAR rays. 
With no more than ten training epochs, our proposed PC-NeRF performs well even when the frame sparsity is 90\,\%. Considering the training time, memory consumption, and inference results for these scenes, a frame sparsity of 67\,\% emerges as a crucial threshold.
Therefore, we subsequently perform additional tests on the KITTI and MaiCity datasets, as depicted in rows 26-31 of Tab.~\ref{tab:Lesser}, to further verify PC-NeRF's performance at this crucial threshold. 
By employing only one of every three LiDAR frames (frame sparsity at 67\,\%) for training, our proposed PC-NeRF maintains robust reliability on these two datasets with just one epoch training. This robust reliability is because our PC-NeRF model concurrently optimizes scene-level, segment-level, and point-level scene representations, capable of utilizing the available point cloud information as efficiently as possible.

Additionally, we report the 3D scene reconstruction outcomes for the experimental groups of Tab.~\ref{tab:Lesser} in Tab.~\ref{tab2} for the same frame sparsities. 
Tab.~\ref{tab2} shows that our proposed PC-NeRF achieves higher 3D scene reconstruction accuracy than NeRF-LOAM on the KITTI and MaiCity datasets at different frame sparsity. Therefore, our proposed PC-NeRF effectively tackles the challenge posed by sparse LiDAR frames, improving the accuracy of 3D reconstruction using sparse LiDAR frames.

\begin{table*}[!t]
\centering
\renewcommand\arraystretch{1.4}		
\caption{Novel LiDAR view synthesis and single-frame 3D reconstruction of PC-NeRF using sparse LiDAR frames on the MaiCity and KITTI datasets (type of depth inference for PC-NeRF: two-step, bolded: best results when training one epoch)}	
{\fontsize{7.35}{8}\selectfont 				
	\begin{tabular}{ccccccccccc}
		\cline{1-11}	 
		\midrule		
		\multicolumn{2}{c}{\begin{tabular}[c]{@{}c@{}}Scene/Dataset\end{tabular}} &
		\multicolumn{1}{c}{\begin{tabular}[c]{@{}c@{}}Training\\set proportion\end{tabular}} &		
		\multicolumn{1}{c}{\begin{tabular}[c]{@{}c@{}}Frame\\sparsity\,[\%]\end{tabular}} & 		
		\multicolumn{1}{c}{\begin{tabular}[c]{@{}c@{}}Train\\epoch\end{tabular}} & 
		\multicolumn{1}{c}{\begin{tabular}[c]{@{}c@{}}Train time/\\epoch\,$[$min$]$\end{tabular}} & \multicolumn{1}{c}{\begin{tabular}[c]{@{}c@{}}Memory\\consumption\end{tabular}} & 
		\multicolumn{1}{c}{\downarrowvaluec{\begin{tabular}[c]{@{}c@{}}Dep.\\Err.\,[$\mathrm{m}$]\end{tabular}}}&
		\multicolumn{1}{c}{\uparrowvaluec{\begin{tabular}[c]{@{}c@{}}Dep. Acc@\\0.2m\,[\%]\end{tabular}}}&
		\multicolumn{1}{c}{\downarrowvalued{\begin{tabular}[c]{@{}c@{}}CD \\$[$m$]$\end{tabular}}}	& 
		\multicolumn{1}{c}{\uparrowvalueb{\begin{tabular}[c]{@{}c@{}}F@\\0.2m\end{tabular}}} \\ 
		\cline{1-11}	 
		\midrule			
		\multicolumn{1}{c}{\multirow{24}[0]{*}{Scene}}&\multicolumn{1}{c}{\multirow{8}[0]{*}{\begin{tabular}[c]{@{}c@{}}MaiCity\\00 sequence\\1-50 frame\end{tabular}}} 
		& 4/5 & 20 & 1 & 90.0  & 12.1\,MB\,+\,275.1\,KB & 0.303  & 88.956  & 0.172 & 0.955    \\
		& & 3/4 & 25 & 1 & 75.7  & 12.1\,MB\,+\,274.1\,KB	& 0.287  & 89.159  & 0.166 & 0.957 \\	
		& & 2/3 & 33 & 1 & 66.3  & 12.1MB\,+\,273.0\,KB   & 0.328  & 88.343  & 0.192 & 0.946   \\
		& & 1/2 & 50 & 1 & 56.1  & 12.1\,MB\,+\,268.9\,KB	& 0.245  & 88.810  & 0.143 & 0.953  \\	
		& & 1/3 & 67 & 1 & 38.0  & 12.1\,MB\,+\,258.7\,KB & 0.189  & 90.168  & 0.109 & 0.961  \\
		& & \textbf{1/4} & \textbf{75} & \textbf{1} & 31.0  & 12.1\,MB\,+\,248.0\,KB &\textbf{0.160}& \textbf{91.219} & \textbf{0.097} & \textbf{0.969} \\	
		\cline{3-11}		
		& & 1/5 & 80 & 5 & 22.4  & 12.1\,MB\,+\,231.2\,KB & 0.177  & 90.283 & 0.106  & 0.962    \\
		& & 1/10& 90 & 5 & 11.3  & 12.1\,MB\,+\,222.3\,KB & 0.163  & 88.747 & 0.108  & 0.946   \\
		\cline{2-11}	 		
		& \multicolumn{1}{c}{\multirow{8}[0]{*}{\begin{tabular}[c]{@{}c@{}}KITTI\\00 sequence\\1151-1200 frame\end{tabular}}} 
		& 4/5 & 20 & 1 & 76.0  & 12.1\,MB\,+\,736.1\,KB & 0.488  & 66.654  & 0.224 & 0.891    \\
		& & 3/4 & 25 & 1 & 55.0  & 12.1\,MB\,+\,734.8\,KB & 0.443  & 69.213  & 0.206 & 0.900  \\			
		& & 2/3 & 33 & 1 & 49.0  & 12.1\,MB\,+\,735.0\,KB & 0.447  & 69.547  & 0.206 & 0.900    \\
		& & 1/2 & 50 & 1 & 35.6  & 12.1\,MB\,+\,702.8\,KB & 0.461 & 68.293 & 0.213 & 0.898 	\\		
		& & \textbf{1/3} & \textbf{67} & \textbf{1} & 24.5  & 12.1\,MB\,+\,676.3\,KB & \textbf{0.439} & \textbf{72.185}  & \textbf{0.197} & \textbf{0.909}  \\
		\cline{3-11}	
		& & 1/4 & 75 & 2 & 24.0  & 12.1\,MB\,+\,662.2\,KB & 0.433 & 71.963 & 0.202 & 0.906  \\			
		& & 1/5 & 80 & 5 & 14.9  & 12.1\,MB\,+\,621.3\,KB & 0.423 & 72.895 & 0.197 & 0.908   \\
		& & 1/10& 90 & 10& 7.3   & 12.1\,MB\,+\,550.2\,KB & 0.412 & 73.620 & 0.197 & 0.904  \\
		\cline{2-11}	 		
		& \multicolumn{1}{c}{\multirow{8}[0]{*}{\begin{tabular}[c]{@{}c@{}}KITTI\\06 sequence\\436-485 frame\end{tabular}}} 
		& 4/5 & 20 & 1 & 68.0 & 12.1\,MB\,+\,693.7\,KB & 0.350 & 72.420  & 0.209  & 0.894  \\
		& & 3/4 & 25 & 1 & 64.4 & 12.1\,MB\,+\,686.6\,KB & 0.337 & 73.829 & 0.206 & 0.898  \\			
		& & 2/3 & 33 & 1 & 44.6 & 12.1\,MB\,+\,687.6\,KB & \textbf{0.330} & 74.303 & 0.204 & \textbf{0.902}  \\
		& & 1/2 & 50 & 1 & 32.8 & 12.1\,MB\,+\,653.2\,KB & 0.332  & \textbf{74.564} & \textbf{0.203} & 0.901  \\	
		& & \textbf{1/3} & \textbf{67} & \textbf{1} & 22.4 & 12.1\,MB\,+\,639.1\,KB & 0.393  & 74.556   & 0.216  & 0.898  \\
		\cline{3-11}		
		& & 1/4 & 75 & 2  & 22.0 & 12.1\,MB\,+\,615.1\,KB & 0.312 & 75.942 & 0.196 & 0.905 \\					
		& & 1/5 & 80 & 10 & 12.3 & 12.1\,MB\,+\,589.5\,KB & 0.323 & 74.808 & 0.200 & 0.905   \\
		& & 1/10& 90 & 10 & 7.6  & 12.1\,MB\,+\,505.5\,KB & 0.311 & 75.424 & 0.196 & 0.902  \\
		\cline{1-11}	 
		\midrule			
		\multicolumn{1}{c}{\multirow{6}[0]{*}{Dataset}} & \multicolumn{1}{c}{\multirow{3}[0]{*}{\begin{tabular}[c]{@{}c@{}}MaiCity 00-01\\sequence\end{tabular}}} 
		& 4/5 & 20  & 1  & 92.0  & 12.1\,MB\,+\,300.7\,KB & 0.347  & 87.877  & 0.179  & 0.945   \\
		& & \textbf{1/3} & \textbf{67}  & \textbf{1}  & 36.0  & 12.1\,MB\,+\,288.4\,KB & 0.237  & 89.287  & 0.123  & 0.954  \\			
		& & 1/10 & 90 & 10 & 11.5  & 12.1\,MB\,+\,249.6\,KB & 0.238 & 87.967 & 0.134 & 0.943 \\
		\cline{2-11}
		& \multicolumn{1}{c}{\multirow{3}[0]{*}{\begin{tabular}[c]{@{}c@{}}KITTI 00-10\\sequence\end{tabular}}} 
		& 4/5  & 20 & 1  & 76.0  & 12.1\,MB\,+\,963.7\,KB & 0.592  & 59.149  & 0.244  & 0.863   \\
		& & \textbf{1/3}  & \textbf{67} & \textbf{1}  & 33.0  & 12.1\,MB\,+\,865.4\,KB & 0.566  & 62.412  & 0.233  & 0.874 	 \\			
		& & 1/10 & 90 & 10 & 9.1   & 12.1\,MB\,+\,657.0\,KB & 0.516  & 64.425  & 0.223  & 0.881  \\
		\cline{1-11}	 
		\midrule			
	\end{tabular}%
}
\label{tab:Lesser}%
\end{table*}%

\subsection{Ablation Study}\label{sec:Ablation}
To validate the effectiveness of our method's components, we conduct ablation experiments probing various aspects of the proposed PC-NeRF. These experiments are tailored to scrutinize the individual components' functionality rather than determining the optimal parameter set. We conduct extensive ablation tests and select the results on the KITTI 00 sequence 1151-1200 frame scene and the KITTI 01 sequence 1001-1050 frame scene for analysis. We train for one epoch and employ the two-step depth inference method for depth inference.

\textbf{Effect of parent NeRF depth loss}: 
As shown in Tab.~\ref{tab:parent NeRF depth loss}, on the KITTI 00 sequence 1151-1200 frame (frame sparsity = 20\,\%) scene, increasing $\lambda_{\mathrm{pd}}$ appropriately improves the accuracy of novel LiDAR view synthesis and single-frame 3D reconstruction. Additionally, compared to $\lambda_{\mathrm{cf}} = 10^6$ and $\lambda_{\mathrm{cd}} = 10^5$, the lower value of $\lambda_{\mathrm{pd}}$ = 100 is the ideal choice. When increasing the frame sparsity to 67\,\%, i.e., drastically reducing the number of laser points for training the PC-NeRF model, maintaining $\lambda_{\mathrm{pd}}$ to a small value (e.g., 0 and 1) achieves a relatively pleasing result on the KITTI 00 sequence 1151-1200 frame scene. This relatively pleasing result is because the parent NeRF free loss proposed in Sec.~\ref{sec:LiDAR loss} is used to supervise the volumetric distribution over the entire scene space, relying on massive actual LiDAR points to obtain the volumetric distribution within the entire scene space. Therefore, to tackle the sparsity of the LiDAR frames, it is not appropriate to set too larger proportion of the parent NeRF free in the total loss. Besides, with $\lambda_{\mathrm{pd}}$ = 1, $\lambda_{\mathrm{cf}} = 10^6$, and $\lambda_{\mathrm{cd}} = 10^5$, we have gotten considerable results with one-epoch training on 13 sequences from MaiCity and KITTI datasets in Sec.~\ref{sec:Evaluating} and Sec.~\ref{sec:Lesser point clouds for 3D reconstruction}. In summary, in order to tackle the scene complexity and effectively utilize the sparse LiDAR frames, we set $\lambda_{\mathrm{pd}}$ to 1 to ensure that our proposed PC-NeRF achieves a relatively high novel LiDAR view synthesis and single-frame 3D reconstruction accuracy in different scenes through fast training.

\textbf{Effect of child NeRF free loss and child NeRF depth loss}: In our proposed PC-NeRF, the child NeRF free loss controlled by $\lambda_{\mathrm{cf}}$ optimizes the scene-level and segment-level environmental representation. As shown in rows 2 to 6 of Tab.~\ref{tab:child_free_depth}, the child NeRF free loss helps improve novel LiDAR view synthesis and single-frame 3D reconstruction accuracy within a specific range of $\lambda_{\mathrm{cf}}$ variation. However, too high or too low $\lambda_{\mathrm{cf}}$ reduces the accuracy. Based on child NeRF free loss, child NeRF depth loss controlled by $\lambda_{\mathrm{cd}}$ and $\lambda_{\mathrm{in}}$ is used to further optimize the point-level and segment-level scene representations. $\lambda_{\mathrm{cd}}$ and $\lambda_{\mathrm{in}}$ are independent, so many pairings of $\lambda_{\mathrm{cd}}$ and $\lambda_{\mathrm{in}}$ need to be experimentally verified. Here, we simplify the pairing by taking $\lambda_{\mathrm{cd}} = \lambda_{\mathrm{cf}} \times \lambda_{\mathrm{in}}$. As shown in rows 7 to 10 of Tab.~\ref{tab:child_free_depth}, unfortunately, adding child NeRF depth loss does not improve the novel LiDAR view synthesis and 3D reconstruction accuracy. So, we further explore the smooth transition between child NeRF free loss and child NeRF depth loss in the following paragraph.

\textbf{Effect of smooth transition between child NeRF free loss and child NeRF depth loss}: As shown in rows 11 to 16 of Tab.~\ref{tab:child_free_depth}, enlarging the smooth transition interval $\gamma$ between child NeRF free loss and child NeRF depth loss improves the novel LiDAR view synthesis and single-frame 3D reconstruction accuracy. At the same time, a sizeable smooth transition interval decreases the accuracy. The accuracy decreases because, when the smoothing transition interval is too large, the child NeRF depth loss needs to supervise almost the sampling points on the entire LiDAR ray and cannot effectively supervise the sampling points around the child NeRF near and far bounds, i.e., around the real objects.

\textbf{Effect of two-step depth inference}: In Fig.~\ref{Inference-effects} and Tab.~\ref{tab:Parent-child NeRF Inference Effect}, two-step depth inference often outperforms one-step depth inference. Moreover, the extensive experiments in Sec.~\ref{sec:Evaluating}, Sec.~\ref{sec:Lesser point clouds for 3D reconstruction}, and Sec.~\ref{sec:Ablation} demonstrate that our proposed PC-NeRF training and two-step depth inference methods are highly compatible and consistently robust.

\newcommand{\uparrowvaluef}[1]{\makebox[0pt][l]{#1}\hspace{1.15cm}\rotatebox[origin=c]{0}{$\uparrow$}}
\newcommand{\downarrowvalueh}[1]{\makebox[0pt][l]{#1}\hspace{0.8cm}\rotatebox[origin=c]{180}{$\uparrow$}}
\begin{table}[!t]
\centering
\renewcommand\arraystretch{1.4}		
\caption{Ablation study of parent NeRF depth loss}	
{\fontsize{7.0}{8}\selectfont 
	\begin{tabular}{cccccc}
		\cline{1-6}
		\midrule			
		\multicolumn{1}{c}{\begin{tabular}[c]{@{}c@{}}Scene\end{tabular}} &		
		\multicolumn{1}{c}{$\lambda_{\mathrm{pd}}$} & 
		\multicolumn{1}{c}{\downarrowvalueh{\begin{tabular}[c]{@{}c@{}}Dep.\\Err.\,[$\mathrm{m}$]\end{tabular}}}&
		\multicolumn{1}{c}{\uparrowvaluef{\begin{tabular}[c]{@{}c@{}}Dep. Acc@\\0.2m\,[\%]\end{tabular}}}&
		\multicolumn{1}{c}{\downarrowvalued{\begin{tabular}[c]{@{}c@{}}CD\\$[$m$]$\end{tabular}}}	& 
		\multicolumn{1}{c}{\uparrowvalueb{\begin{tabular}[c]{@{}c@{}}F@\\0.2m\end{tabular}}} \\
		\cline{1-6}
		\midrule
		\multicolumn{1}{c}{\multirow{7}[0]{*}{\begin{tabular}[c]{@{}c@{}}KITTI\\00 sequence\\1151-1200\\frame\\(frame sparsity\\= 20\,\%)\end{tabular}}} & 0     & 0.4892  & 66.6861  & 0.2243  & 0.8910  \\
		&1 (our)    & 0.4884  & 66.6541  & 0.2239  & 0.8908  \\
		&10    & 0.4895  & 66.6587  & 0.2241  & 0.8908  \\
		&100   & \textbf{0.4675}  & 69.5239  & \textbf{0.2077}  & \textbf{0.8993}  \\
		&1.0e3  & 0.4683  & 69.7551  & 0.2089  & 0.8989  \\
		&1.0e4 & 0.4729  & \textbf{69.9102}  & 0.2079  & 0.8992  \\
		&1.0e5 & 0.5307  & 67.6254  & 0.2240  & 0.8931  \\	
		\cline{1-6}	
		\multicolumn{1}{c}{\multirow{7}[0]{*}{\begin{tabular}[c]{@{}c@{}}KITTI\\00 sequence\\1151-1200\\frame\\(frame sparsity\\= 67\,\%)\end{tabular}}} & 0     & \textbf{0.4384}  & 72.1457  & \textbf{0.1966}  & 0.9086  \\
		& 1 (our)     & 0.4390  & 72.1848  & 0.1967  & 0.9086  \\
		& 10    & 0.4400  & 72.1695  & 0.1969  & 0.9085  \\
		& 100   & 0.4454  & 72.1620  & 0.1988  & 0.9078  \\
		& 1.0e3  & 0.4540  & 72.4099  & 0.2005  & 0.9062  \\
		& 1.0e4 & 0.4443  & \textbf{72.5878}  & 0.1969  & \textbf{0.9091}  \\
		& 1.0e5 & 0.4977  & 70.7471  & 0.2112  & 0.9035  \\		
		\cline{1-6}									
		\multicolumn{1}{c}{\multirow{7}[0]{*}{\begin{tabular}[c]{@{}c@{}}KITTI\\01 sequence\\1001-1050\\frame\\(frame sparsity\\= 20\,\%)\end{tabular}}} & 0     & 1.3424  & 39.6964  & 0.3521  & 0.7422  \\
		&1 (our)     & 1.3358  & 39.6509  & \textbf{0.3501}  & \textbf{0.7439}  \\
		&10    & 1.3435  & 39.6998  & 0.3526  & 0.7417  \\
		&100   & 1.3490  & 39.7598  & 0.3557  & 0.7384  \\
		&1.0e3  & 1.3259  & \textbf{39.9901}  & 0.3567  & 0.7361  \\
		&1.0e4 & 1.2874  & 38.8882  & 0.3596  & 0.7266  \\
		&1.0e5 & \textbf{1.2785}  & 38.6314  & 0.3580  & 0.7268  \\
		\cline{1-6}	
		\multicolumn{5}{l}{Note: $\lambda_{\mathrm{cf}} = 1.0e6, \lambda_{\mathrm{cd}} = 1.0e5, \lambda_{\mathrm{in}} = 0.1,\gamma = 2\,\mathrm{m}$} \\
		\cline{1-6}
		\midrule			
	\end{tabular}%
}
\label{tab:parent NeRF depth loss}%
\end{table}%

\newcommand{\downarrowvaluee}[1]{\makebox[0pt][l]{#1}\hspace{0.7cm}\rotatebox[origin=c]{180}{$\uparrow$}}
\newcommand{\uparrowvalued}[1]{\makebox[0pt][l]{#1}\hspace{1.0cm}\rotatebox[origin=c]{0}{$\uparrow$}}
\newcommand{\downarrowvaluef}[1]{\makebox[0pt][l]{#1}\hspace{0.3cm}\rotatebox[origin=c]{180}{$\uparrow$}}
\begin{table}[!t]
\centering
\renewcommand\arraystretch{1.4}	
\caption{Ablation study of child NeRF free loss and child NeRF depth loss on the KITTI 00 sequence 1151-1200 frame scene ($\lambda_{\mathrm{pd}}$: 1, frame sparsity: 20\,\%.)}			
{\fontsize{6.5}{8}\selectfont 
	\begin{tabular}{cccccccc}
		\cline{1-8}		
		\midrule
		\multicolumn{1}{c}{$\lambda_{\mathrm{cf}}$} &
		\multicolumn{1}{c}{$\lambda_{\mathrm{cd}}$} & 
		\multicolumn{1}{c}{$\lambda_{\mathrm{in}}$} &	
		\multicolumn{1}{c|}{$\gamma\,[\mathrm{m}]$} &					\multicolumn{1}{c}{\downarrowvaluee{\begin{tabular}[c]{@{}c@{}}Dep.\\Err.\,[$\mathrm{m}$]\end{tabular}}}&
		\multicolumn{1}{c}{\uparrowvalued{\begin{tabular}[c]{@{}c@{}}Dep. Acc@\\0.2m\,[\%]\end{tabular}}}&
		\multicolumn{1}{c}{\downarrowvaluef{\begin{tabular}[c]{@{}c@{}}CD\\$[$m$]$\end{tabular}}}	& 
		\multicolumn{1}{c}{\uparrowvalueb{\begin{tabular}[c]{@{}c@{}}F@\\0.2m\end{tabular}}} \\
		\cline{1-8}		
		\midrule		
		0 & \multirow{5}[0]{*}{0} & \multirow{5}[0]{*}{0} &  \multicolumn{1}{c|}{\multirow{5}[0]{*}{0}}   & 0.511  & 65.232  & 0.220  & 0.890  \\
		1  & &  &\multicolumn{1}{c|}{}  & 0.507  & 66.129  & 0.217  & 0.894  \\
		1.0e3 & & & \multicolumn{1}{c|}{}  & 0.472  & \textbf{69.548}  & \textbf{0.209}  & \textbf{0.899}  \\
		1.0e6 & & & \multicolumn{1}{c|}{}  & \textbf{0.465}  & 67.109  & 0.219  & 0.894  \\
		1.0e9 & & & \multicolumn{1}{c|}{}  & 0.466  & 67.093  & 0.219  & 0.893  \\		
		\cline{1-8}						
		\multirow{4}[0]{*}{1.0e6} & 2.5e4 & 0.025 &  \multicolumn{1}{c|}{\multirow{4}[0]{*}{0}} & \textbf{0.480}  & \textbf{67.001}  & \textbf{0.222}  & \textbf{0.891}  \\
		& 5.0e4 & 0.05 & \multicolumn{1}{c|}{} & 0.494  & 66.819  & 0.224  & 0.890  \\
		& 1.0e5 & 0.1  & \multicolumn{1}{c|}{} & 0.509  & 66.548  & 0.228  & 0.887  \\
		& 2.0e5 & 0.2  & \multicolumn{1}{c|}{} & 0.538  & 65.988  & 0.234  & 0.884  \\
		\cline{1-8}		
		\multirow{6}[0]{*}{1.0e6} & \multirow{6}[0]{*}{1.0e5} & \multicolumn{1}{c}{\multirow{6}[0]{*}{0.1}} & \multicolumn{1}{c|}{0.5} &  0.511 & 66.430 & 0.228 & 0.887 \\
		&  &  & \multicolumn{1}{c|}{1} & 0.513 & 65.970 & 0.229 & 0.887 \\
		&  &  & \multicolumn{1}{c|}{2 (our)}  & 0.488 & 66.654 & 0.224 & 0.891 \\
		&  &  & \multicolumn{1}{c|}{3} & \textbf{0.447} & 68.697 & 0.207 & 0.899 \\
		&  &  & \multicolumn{1}{c|}{5} &   0.452 & \textbf{69.344} & \textbf{0.206} & \textbf{0.899} \\
		&  &  & \multicolumn{1}{c|}{10} &  0.468 & 70.269 & 0.209 & 0.899 \\
		\cline{1-8}			
		\midrule		
	\end{tabular}%
}
\label{tab:child_free_depth}%
\end{table}%

\section{Conclusion}
This paper proposes a parent-child neural radiance fields (PC-NeRF) framework for large-scale 3D scene reconstruction and novel LiDAR view synthesis optimized for efficiently utilizing temporally sparse LiDAR frames in outdoor autonomous driving.
PC-NeRF proposes a hierarchical spatial partitioning approach to divide the autonomous vehicle driving environment into large blocks, referred to as parent NeRFs, and subsequently subdivide each block into geometric segments, represented by child NeRFs. A parent NeRF shares a network with the child NeRFs within it. 
Leveraging the hierarchical spatial partitioning approach, PC-NeRF introduces a comprehensive multi-level scene representation. This representation is crafted to collectively optimize scene-level, segment-level, and point-level features, enabling efficient utilization of sparse LiDAR frames.
Besides, we propose a two-step depth inference method to realize segment-to-point inference. Our proposed PC-NeRF is validated with extensive experiments to achieve high-precision novel LiDAR view synthesis and 3D reconstruction in large-scale scenes. 
Furthermore, PC-NeRF demonstrates significant deployment potential, achieving notable accuracy in novel LiDAR view synthesis and 3D reconstruction with just one epoch of training on most test scenes from the KITTI and MaiCity datasets.
Most importantly, PC-NeRF can tackle practical situations with temporally sparse LiDAR frames.
Our future work will further explore the potential of PC-NeRF in object detection and localization for autonomous driving.

\bibliographystyle{IEEEtran}      
\bibliography{reference}                        

\end{document}